\documentclass{article}

\usepackage{microtype}
\usepackage{graphicx}
\usepackage{subcaption}
\usepackage{booktabs} % for professional tables

\usepackage{hyperref}

\usepackage[preprint]{icml2026}

\usepackage{amsmath}
\usepackage{amssymb}
\usepackage{mathtools}
\usepackage{amsthm}

\newcommand{\framework}{\textsc{AltTS}}
\usepackage{url}
\usepackage{multirow}
\usepackage{enumitem}
\usepackage{amsmath,amsfonts,bm}
\usepackage{algorithm}
\usepackage{algorithmic}
\usepackage[inkscapelatex=false]{svg}

\def\eqref#1{equation~\ref{#1}}
\def\1{\bm{1}}

\def\rvr{{\mathbf{r}}}

\def\rvx{{\mathbf{x}}}
\def\rvy{{\mathbf{y}}}

\def\rmA{{\mathbf{A}}}

\def\rmF{{\mathbf{F}}}

\def\rmK{{\mathbf{K}}}

\def\rmM{{\mathbf{M}}}

\def\rmQ{{\mathbf{Q}}}

\def\rmV{{\mathbf{V}}}

\def\rmX{{\mathbf{X}}}
\def\rmY{{\mathbf{Y}}}
\def\rmZ{{\mathbf{Z}}}

\DeclareMathAlphabet{\mathsfit}{\encodingdefault}{\sfdefault}{m}{sl}
\SetMathAlphabet{\mathsfit}{bold}{\encodingdefault}{\sfdefault}{bx}{n}

\newcommand{\E}{\mathbb{E}}

\newcommand{\R}{\mathbb{R}}

\newcommand{\Cov}{\mathrm{Cov}}
\DeclareMathOperator*{\argmin}{arg\,min}

\DeclareMathOperator{\Tr}{Tr}

\usepackage[capitalize,noabbrev]{cleveref}

\theoremstyle{plain}
\newtheorem{theorem}{Theorem}[section]
\newtheorem{proposition}[theorem]{Proposition}

\theoremstyle{definition}

\newtheorem{assumption}[theorem]{Assumption}
\theoremstyle{remark}

\usepackage[textsize=tiny]{todonotes}

\icmltitlerunning{AltTS: A Dual-Path Framework with Alternating Optimization for Multivariate Time Series Forecasting}

\begin{document}

\twocolumn[
  \icmltitle{AltTS: A Dual-Path Framework with Alternating Optimization for Multivariate Time Series Forecasting}

  % It is OKAY to include author information, even for blind submissions: the
  % style file will automatically remove it for you unless you've provided
  % the [accepted] option to the icml2026 package.

  % List of affiliations: The first argument should be a (short) identifier you
  % will use later to specify author affiliations Academic affiliations
  % should list Department, University, City, Region, Country Industry
  % affiliations should list Company, City, Region, Country

  % You can specify symbols, otherwise they are numbered in order. Ideally, you
  % should not use this facility. Affiliations will be numbered in order of
  % appearance and this is the preferred way.
  \icmlsetsymbol{equal}{*}

  \begin{icmlauthorlist}
    \icmlauthor{Zhihang Yuan}{equal,uoe}
    \icmlauthor{Zhiyuan Liu}{equal,uc}
    \icmlauthor{Mahesh K. Marina}{uoe}
    %\icmlauthor{}{sch}
    %\icmlauthor{}{sch}
    %\icmlauthor{}{sch}
  \end{icmlauthorlist}

  \icmlaffiliation{uoe}{School of Informatics, University of Edinburgh, UK}
  \icmlaffiliation{uc}{University of Chicago, USA}

  % \icmlcorrespondingauthor{Zhihang Yuan}{zhihang.yuan@ed.ac.uk}
  % \icmlcorrespondingauthor{Zhiyuan Liu}{zhiyuanliu@uchicago.edu}
  \icmlcorrespondingauthor{Mahesh K. Marina}{mahesh@ed.ac.uk}

  % You may provide any keywords that you find helpful for describing your
  % paper; these are used to populate the "keywords" metadata in the PDF but
  % will not be shown in the document
  \icmlkeywords{Machine Learning, ICML}

  \vskip 0.3in
]

% this must go after the closing bracket ] following \twocolumn[ ...

% This command actually creates the footnote in the first column listing the
% affiliations and the copyright notice. The command takes one argument, which
% is text to display at the start of the footnote. The \icmlEqualContribution
% command is standard text for equal contribution. Remove it (just {}) if you
% do not need this facility.

% Use ONE of the following lines. DO NOT remove the command.
% If you have no special notice, KEEP empty braces:
% \printAffiliationsAndNotice{}  % no special notice (required even if empty)
% Or, if applicable, use the standard equal contribution text:
\printAffiliationsAndNotice{\icmlEqualContribution}

\begin{abstract}
    Multivariate time series forecasting involves two qualitatively distinct factors: (i) stable within-series autoregressive (AR) dynamics, and (ii) intermittent cross-dimension interactions that can become spurious over long horizons.
    We argue that fitting a single model to capture both effects creates an optimization conflict: the high-variance updates needed for cross-dimension modeling can corrupt the gradients that support autoregression, resulting in brittle training and degraded long-horizon accuracy.
    To address this, we propose \framework, a dual-path framework that explicitly decouples autoregression and cross-relation (CR) modeling.
    In \framework, the AR path is instantiated with a linear predictor, while the CR path uses a Transformer equipped with Cross-Relation Self-Attention (CRSA); the two branches are coordinated via alternating optimization to isolate gradient noise and reduce cross-block interference.
    Extensive experiments on multiple benchmarks show that \framework\ consistently outperforms prior methods, with the most pronounced improvements on long-horizon forecasting.
    Overall, our results suggest that carefully designed optimization strategies, rather than ever more complex architectures, can be a key driver of progress in multivariate time series forecasting.
    The code will be open-sourced upon publication.
\end{abstract}

\section{Introduction}

Time series forecasting aims to estimate future outcomes from past observations. Classical multivariate time series analysis rests on structural and probabilistic assumptions that enable tractable inference. Vector autoregression (VAR) offers a concise baseline for joint stationary dynamics \citep{VAR}, while cointegration theory captures co-movements among non-stationary series \citep{cointegration}. Modern methodologies further provide systematic approaches for long-horizon, high-dimensional time series with diverse cross-time, cross-variable interactions \citep{neweywest, numberoffactors, sparsehigh}. In those statistical methods, modeling cross-variable dependencies is fundamental to understanding the multivariate system.

Building on these foundations, recent deep learning methods extend time series modeling and have demonstrated significant scalability and strong inference ability for long-term time series forecasting (LTSF). Representative lines include CNN- and RNN-based \citep{Lai2017ModelingLA, li2018dcrnn, liu2022scinet}, graph-based \citep{yu2018spatio, shang2021discrete}, and Transformer-based \citep{haoyietal-informer-2021, zhou2022fedformer} models. These models tailor the network structure to the characteristics of time series data, among which techniques such as patching \citep{Yuqietal-2023-PatchTST}, sparse modeling \citep{lin2024sparsetsf} and dependency modeling \citep{zhang2023crossformer, hu2025timefilter} have delivered notable gains across various real-world settings. At the same time, the recent success of MLP-based and linear models \citep{Oreshkin2020N-BEATS:, dlinear, das2023longterm, huang2025timebase} call into question whether increasingly complex architectures are necessary for LTSF, and motivate re-examining how to represent the core properties of multivariate time series for effective forecasting.

From an optimization perspective, training of modern neural networks operates in a non-convex and often non-smooth regime. Foundational analyses establish the convergence of stochastic first-order methods, while adaptive approaches such as Adam \citep{2015-kingma} have inspired refinements and provably convergent variants like AMSGrad \citep{AMSGrad}. For structured objectives, proximal alternating methods \citep{PALM} and block-coordinate descent (BCD) guarantee monotone descent and convergence, which are attracting growing interest within the deep learning community \citep{gddnn, bcddl}. More broadly, the integration of optimization principles with network design, exemplified by embedding differentiable optimization layers in networks \citep{optnet} and unfolding iterative algorithms for model alignment \citep{luo2020unfolding}, has emerged as a powerful paradigm for taming ill-conditioned learning dynamics. In the context of LTSF, unstable learning dynamics can originate from heterogeneous dependency structures. Figure~\ref{fig:grad-variance-7} reports the variance of gradient for autoregression (AR) and cross-relation (CR) parameters under the joint and the alternating training schedules. Since AR and CR patterns differ in learning difficulty, the two components often require different step sizes to converge. With a shared optimizer in the joint training, the CR block exhibits exploding variance in five of seven datasets, whereas the variance with the AR block remains comparatively low and decays over epochs. The high variance of the CR parameter updates injects non-negligible noise into the entire model. The alternating method mitigates this instability by using separate optimizers for AR and CR, yielding more stable convergence for both paths.

Based on the above observations, we propose \framework, a dual-path framework that decouples dependency modeling and trains the two paths via alternating optimization. In \framework, AR and CR are separately forecasted by two modules. Subsequently, alternating optimization is applied to alleviate the entanglement of seemingly homogeneous AR and CR through cyclic block-wise updates. Experimentally, \framework~achieves state-of-the-art performance in a wide range of time series forecasting tasks with minimal architectural sophistication. Our contributions are as follows:
% \begin{enumerate}
%     \item We generalize dependency modeling for multivariate time series forecasting and discuss the learning dynamics under both channel-independent and channel-dependent regimes. We identify the risk of gradient entanglement in time series forecasting when the true longitudinal is unobservable. 
%     \item We propose \framework, which explicitly decouples per-series autoregression and cross-variable dependency, trained with an alternating optimization scheme. \framework\ achieves competitive performance across various real-world datasets using canonical architectures.
%     \item \framework\ motivates treating the training schedules as a design variable, indicating the potential of integrating optimization principles into the structural design of neural networks.
% \end{enumerate}

\begin{enumerate}
    \item We present \framework, the first deep learning framework that explicitly \emph{decouples} autoregression and cross-variable dependency and coordinates them via \emph{alternating optimization}. 
    Our analysis further reveals the risk of gradient entanglement when these patterns are learned jointly, especially when the true longitudinal structure is unobservable. 
    
    \item We evaluate \framework\ on seven widely used multivariate time series benchmarks against strong linear, Transformer-based, and hybrid baselines. 
    \framework\ consistently achieves competitive or superior performance across datasets and horizons while relying only on canonical architectures, underscoring the effectiveness of optimization-driven design. 
    
    \item Beyond empirical results, \framework\ highlights that \emph{training schedules can be treated as a design variable}, pointing to new opportunities for integrating optimization principles into the structural design of neural networks. 
\end{enumerate}

% Make all panels the same visual size
\newlength{\gvpanelheight}
\setlength{\gvpanelheight}{0.18\textwidth} % tweak if you need larger/smaller panels

\begin{figure*}[t]
  \centering
  % ---------- Top row: Weather / Traffic / Electricity ----------
  \begin{subfigure}[t]{0.32\linewidth}
    \centering
    \includegraphics[height=\gvpanelheight,keepaspectratio]{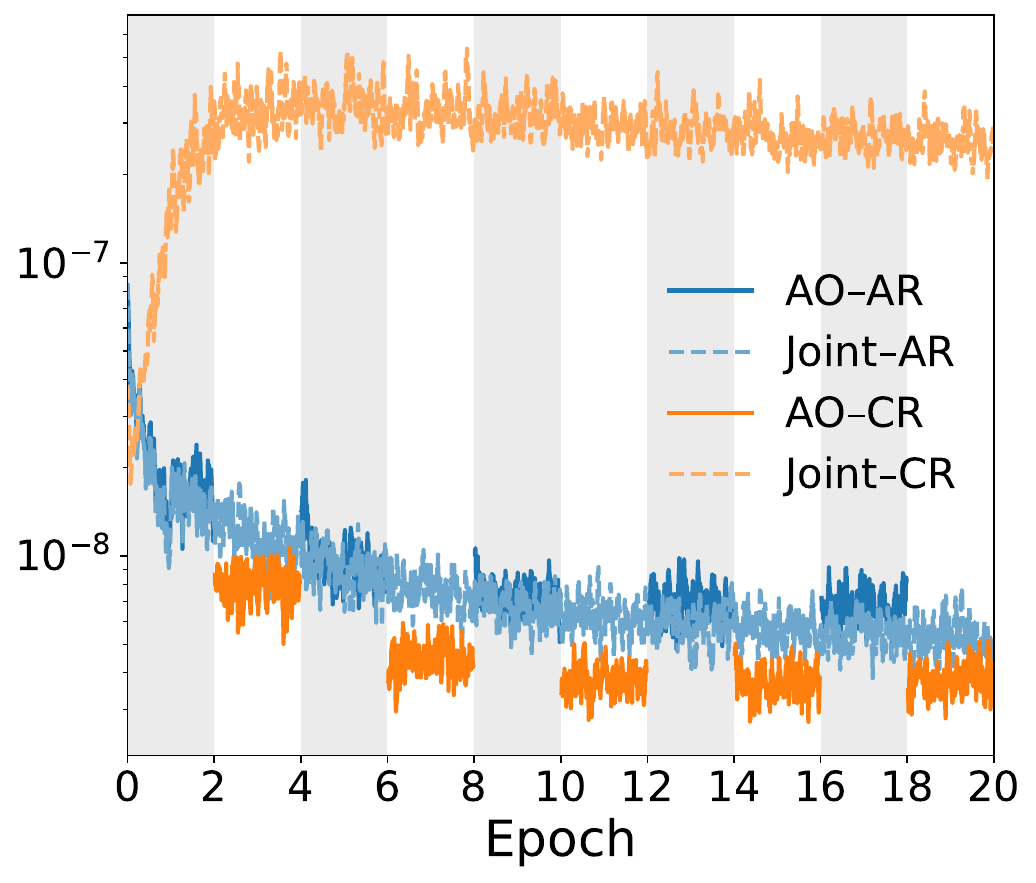}
    \caption{Weather}
    \label{fig:grad-weather}
  \end{subfigure}\hfill
  \begin{subfigure}[t]{0.32\linewidth}
    \centering
    \includegraphics[height=\gvpanelheight,keepaspectratio]{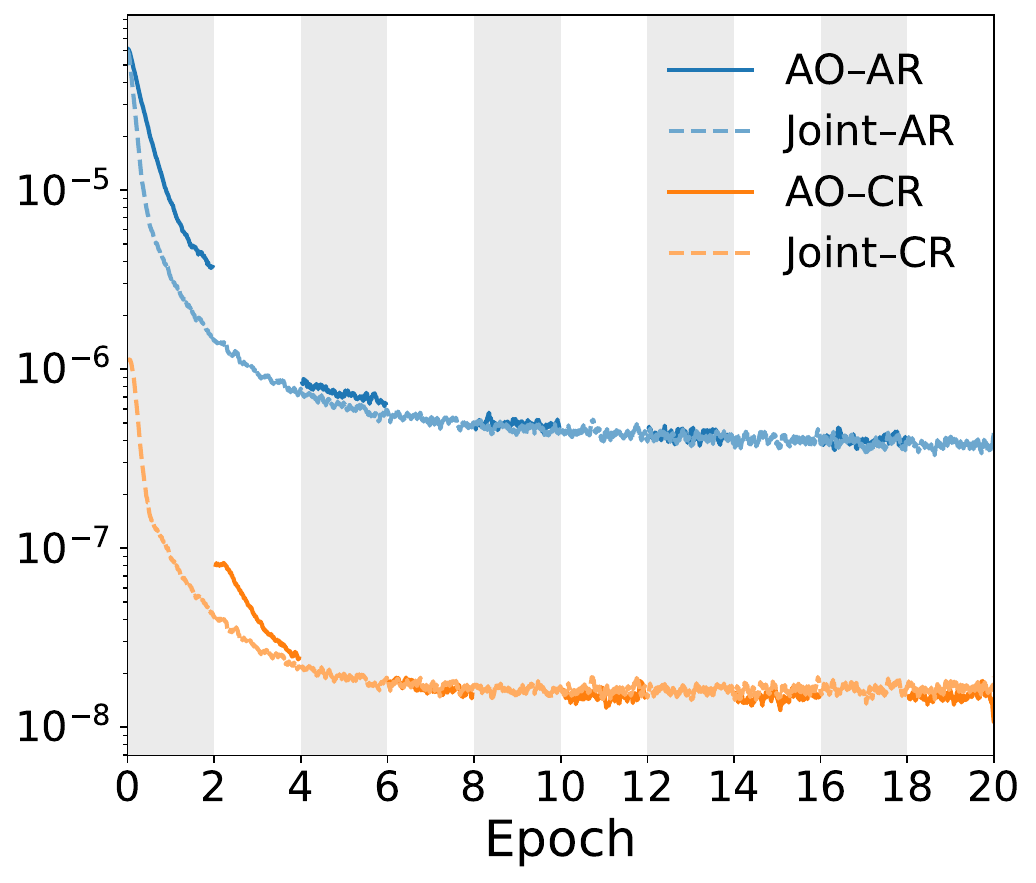}
    \caption{Traffic}
    \label{fig:grad-traffic}
  \end{subfigure}\hfill
  \begin{subfigure}[t]{0.32\linewidth}
    \centering
    \includegraphics[height=\gvpanelheight,keepaspectratio]{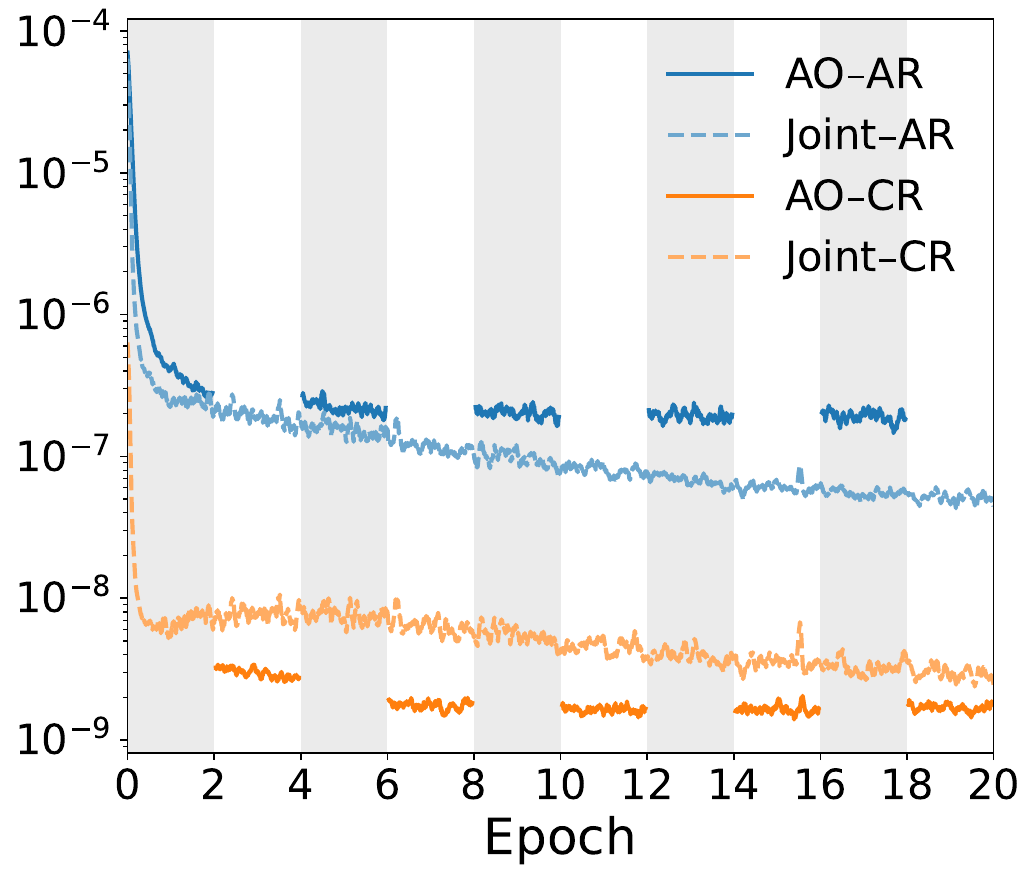}
    \caption{Electricity}
    \label{fig:grad-electricity}
  \end{subfigure}

  \vspace{0.8em}

  % ---------- Bottom row: ETTh1 / ETTh2 / ETTm1 / ETTm2 ----------
  \begin{subfigure}[t]{0.24\linewidth}
    \centering
    \includegraphics[height=\gvpanelheight,keepaspectratio]{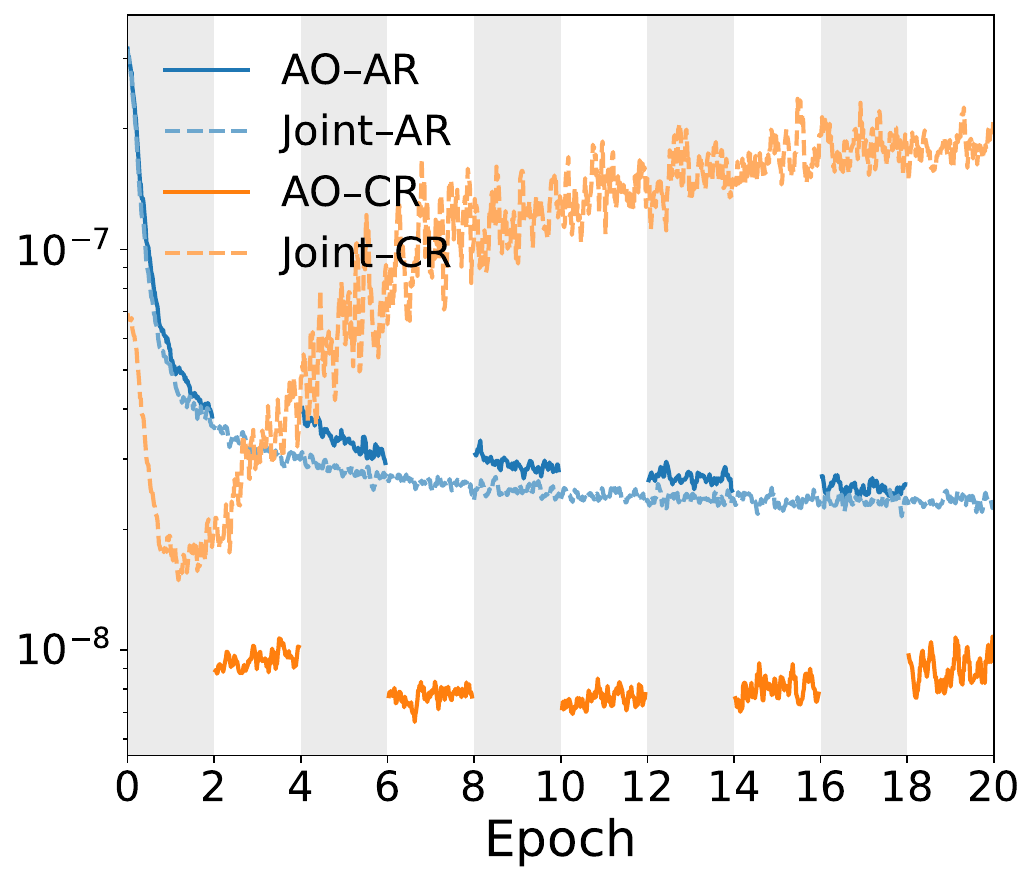}
    \caption{ETTh1}
    \label{fig:grad-etth1}
  \end{subfigure}\hfill
  \begin{subfigure}[t]{0.24\linewidth}
    \centering
    \includegraphics[height=\gvpanelheight,keepaspectratio]{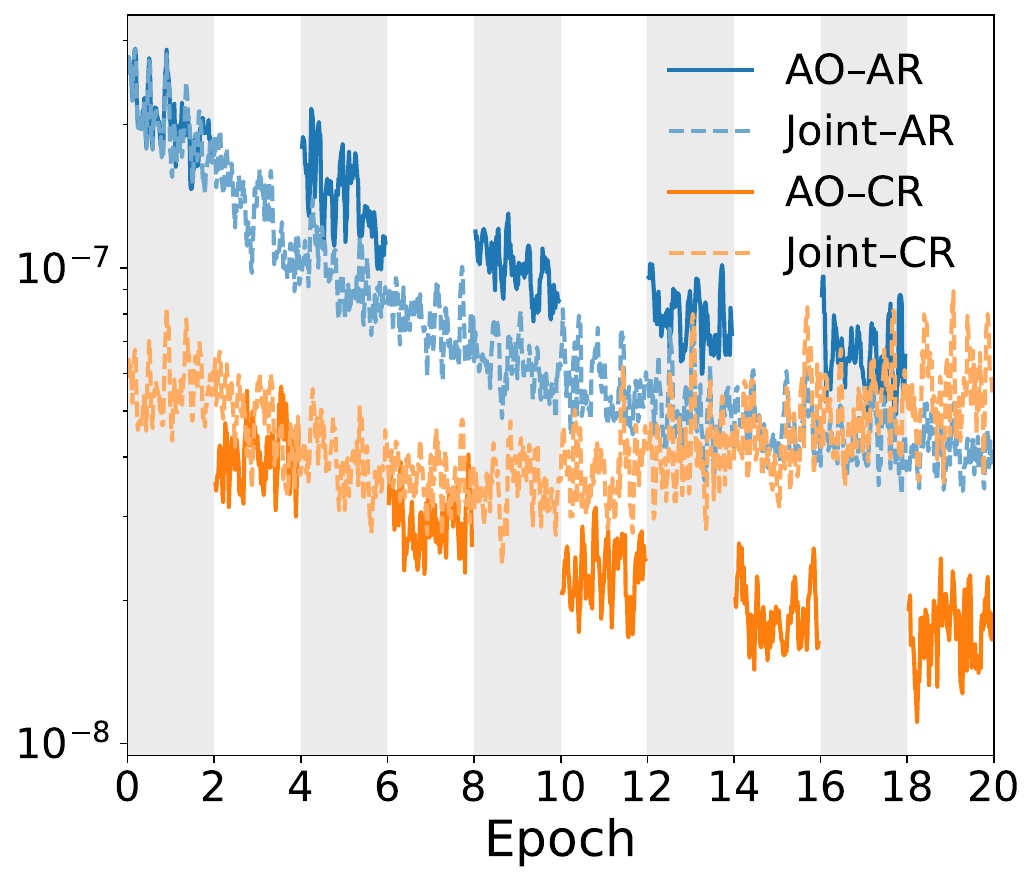}
    \caption{ETTh2}
    \label{fig:grad-etth2}
  \end{subfigure}\hfill
  \begin{subfigure}[t]{0.24\linewidth}
    \centering
    \includegraphics[height=\gvpanelheight,keepaspectratio]{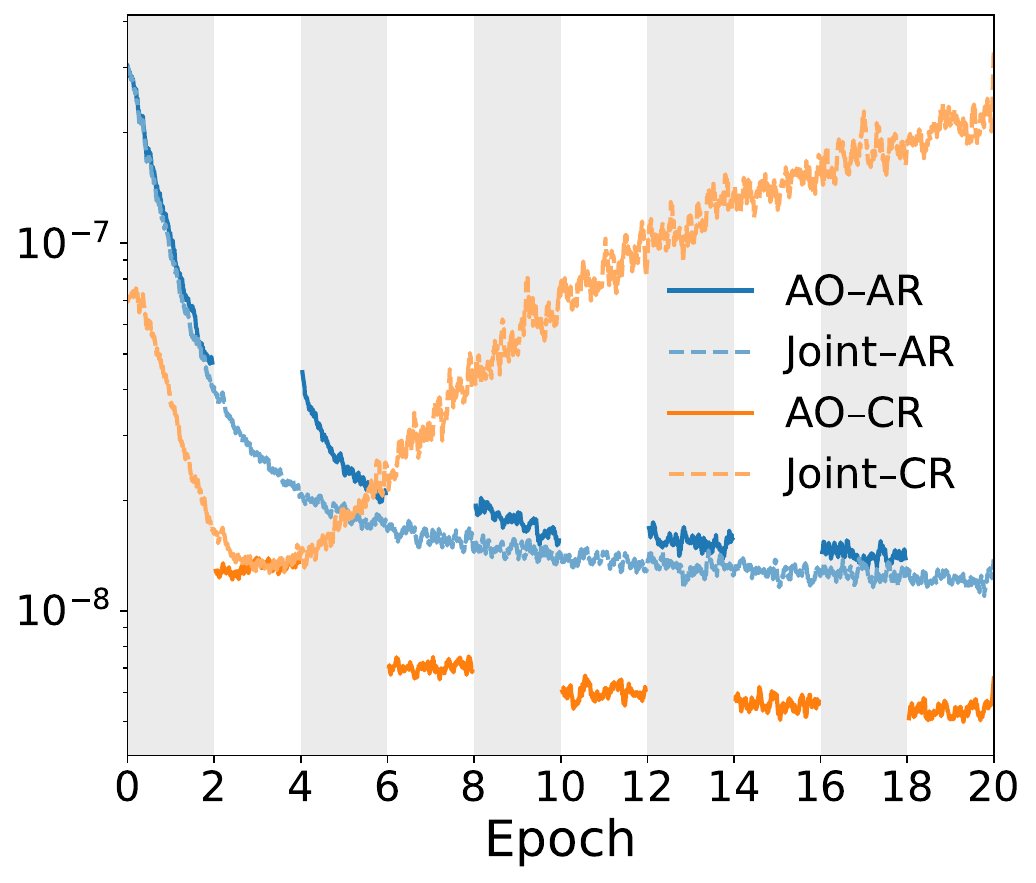}
    \caption{ETTm1}
    \label{fig:grad-ettm1}
  \end{subfigure}\hfill
  \begin{subfigure}[t]{0.24\linewidth}
    \centering
    \includegraphics[height=\gvpanelheight,keepaspectratio]{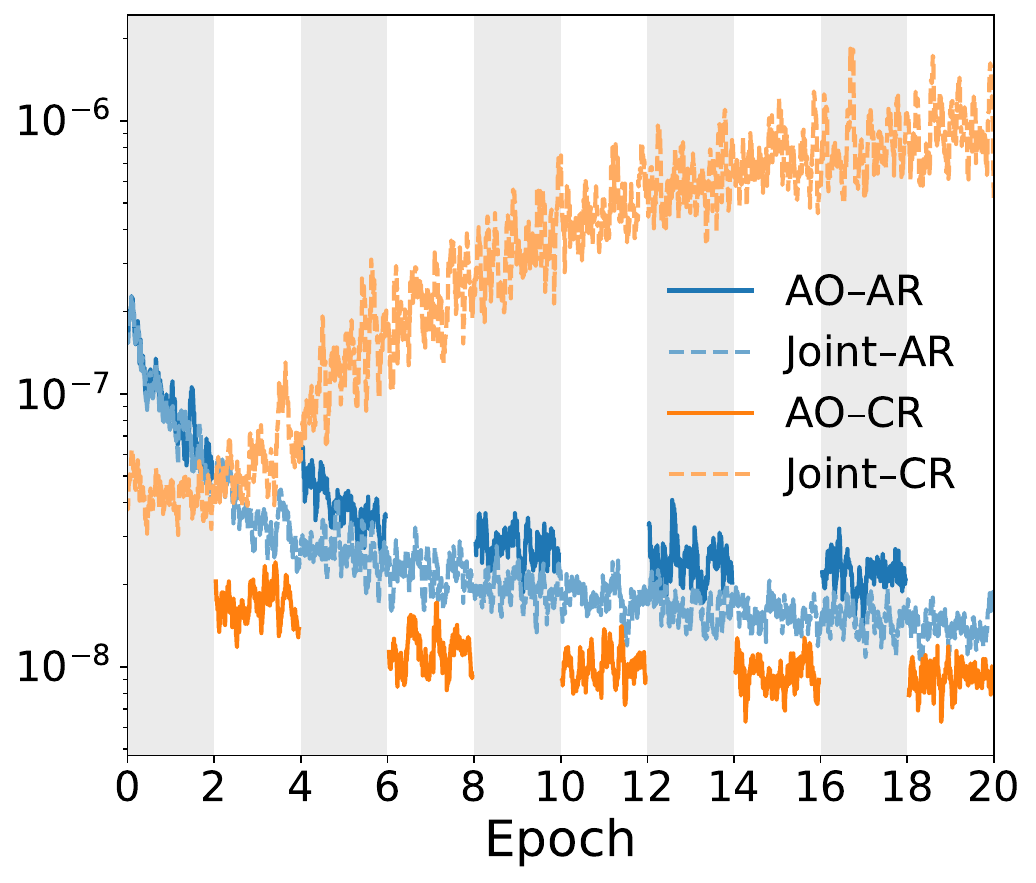} % fixed filename
    \caption{ETTm2}
    \label{fig:grad-ettm2}
  \end{subfigure}

  % \caption{Variance of AR/CR gradients under \emph{alternating} vs. \emph{joint} training across seven datasets (prediction length $=96$). The y-axis shows the log-variance of AR/CR gradients. Higher variance indicates greater training instability, motivating alternating optimization (AO) in \framework.}
  \caption{Variance of AR/CR gradients under \emph{alternating} vs.\ \emph{joint} training on seven datasets (prediction length $=96$). 
  The y-axis is the natural log of the gradient variance statistic. 
  We compute the rolling sample variance of gradients for each parameter over the last $K$ updates; within each branch, we take the parameter-wise sum to yield a scalar for AR and CR in the rolling window, respectively.
  Extended plots for horizons $192/336/720$ are in Appendix~\ref{app:extended-gv}.}

  \label{fig:grad-variance-7}
\end{figure*}

\section{Related Work}

\textbf{Long-Term Time Series Forecasting}\quad
Transformer-based models have been a dominant thread in long-term time series forecasting (LTSF). Many advancements involve adapting the Transformer \citep{AttentionIsAllYouNeed} for LTSF. These models utilize various properties of long-term time series through seasonal-trend decomposition \citep{wu2021autoformer, wang2024timemixer}, frequency analysis \citep{zhou2022fedformer, chen2024pathformer}, and sparse modeling \citep{haoyietal-informer-2021, deformabletst}. Further investigations into the characteristics of time-series data show that simple models, such as linear layers \citep{dlinear, rlinear}, can achieve comparable or superior performance to Transformers. 
More recently, OLinear~\citep{yue2025olinear} further demonstrates that linear forecasters can be highly competitive by operating in an orthogonally transformed domain to alleviate entangled temporal dependencies.
In parallel, there is a growing interest in dependency modeling. The Channel Independent (CI) methods rely solely on the historical information of each series, individually or through a pooled Channel Mix (CM) modeling, thus concentrating on autoregressive patterns. PatchTST \citep{Yuqietal-2023-PatchTST} proposes a CI/CM patching, showing consistent gains on long horizons. Following this path, pure autoregressive models including TimesNet \citep{wu2023timesnet}, SparseTSF \citep{lin2024sparsetsf}, and TimeBase \citep{huang2025timebase} achieve impressive performance, demonstrating the effectiveness of CI modeling. However, the CI/CM modeling overlooks cross-variable dependency, leading to a fundamental deviation of the learned model from the actual data-generating processes. In contrast, the Channel Dependent (CD) methods explicitly target cross-dimension structures. Crossformer \citep{zhang2023crossformer} designs a two-stage attention to capture cross-time and cross-dimension dependencies. iTransformer \citep{liu2023itransformer} tokenizes variables and applies attention to their time embeddings, where attention scores implicitly represent multivariate correlations. To resolve the high variable dimensionality in large datasets, Channel Clustering (CC) is proposed, analogous to patching in time domain modeling. In CC methods, such as DUET \citep{qiu2025duet} and TimeFilter \citep{hu2025timefilter}, heterogeneous variables are grouped together to preserve instantaneous correlations. The two-stage temporal-spatial paradigm is widely adopted in most CD methods. While recent studies also experiment on structures compatible with both CI and CD settings \citep{lin2025petformer}, further research on the integration of autoregressive and cross-dimension modeling is still needed.

\textbf{Alternating Optimization}\quad
Alternating optimization methods attract increasing interest in deep learning, such as Block-Coordinate Descent (BCD) and Alternating Direction Method of Multipliers (ADMM) by \citet{ADMM}. Algorithms that allow general non-smooth and non-convex problems provide theoretical foundations for the application in deep learning scenarios \citep{BSUM, PALM, NIPS2014Parallel}. Meanwhile, the alternating optimization algorithms are employed heuristically in adversarial learning \citep{NIPS2014GAN} and Computer Vision \citep{luo2020unfolding, akbari2023alternating}, where two or more neural networks and objectives are alternately optimized for better alignment. In this paper, we explore the alternating optimization of dependency modeling schemes for LTSF to coordinate autoregressive and cross-dimension patterns while preserving temporal causality.

\section{Methodology}

\subsection{General Structure}
In the multivariate time series forecasting task, let $\rmX_t = [\rvx^{(1)}_t, \dots, \rvx^{(D)}_t] \in \R^{D\times L}$ be the lookback window at step $t$, where each $\rvx^{(i)}_t = [x^{(i)}_{t-L+1}, \dots, x^{(i)}_t]$ represents a historical input sequence of length $L$. $D$ denotes the number of dimensions. The target sequence is $\rmY_{t+1} = [\rvy^{(1)}_{t+1}, \dots, \rvy^{(D)}_{t+1}] \in \R^{D\times H}$, where $\rvy^{(i)}_{t+1} = [x^{(i)}_{t+1}, \dots, x^{(i)}_{t+H}]$ is the realization of the $i$-th sequence from step $t+1$ to $t+H$. For the individual sequences, we omit the subscript $t$ and use $\rvx_i:=\rvx^{(i)}_t,\ \rvy_i:=\rvy^{(i)}_{t+1}$ for short in the following analysis.

In general, to generate the predicted series $\hat{\rmY}_{t+1}$, a transition matrix is used. Let $\rmF=(f_{ij})_{D\times D}$ be a matrix of projections. Each projection $f_{ij}:\R^L\mapsto\R^H$ measures the contribution of $\rvx_j$ to $\rvy_i$. $\rmF$, $f_{ij}$ can be approximated by a neural network or part of a neural network, denoted by $\hat{\rmF}$ and $\hat{f}_{ij}$ respectively. For convenience, we define an ``apply-then-sum'' operator $*$ as
\begin{equation}
    \rmF *\rmX_t := \left(\sum_{j=1}^D f_{1j}(\rvx_j), \dots, \sum_{j=1}^D f_{Dj}(\rvx_j)\right) \in \R^{D\times H}.
\end{equation}
Note that the $*$ operation is linear with respect to the first argument. The transition equation is therefore
\begin{equation}\label{generalform}
    \rmF *\rmX_t + \rmV_{t+1} = \rmY_{t+1},
\end{equation}
where $\rmV_{t+1}$ represents the unpredictable innovations of the time series at step $t+1$ given $\rmX_t$. This general form summarizes CI, CM, and CD methods. Setting all off-diagonal projections to zero, $f_{ij}\equiv 0, i\ne j$ yields the CI methods, where $D$ independent neural networks are used to model the autoregression per series. Further imposing $f_{11}=\dots=f_{DD}$ gives the CM setting, where a single shared neural network generalizes the autoregressive patterns from all series. Allowing non-trivial off-diagonal entries, i.e., at least one $f_{ij},i\ne j$ depends on the input, leads to the CD methods with explicit cross-dimension effects.

However, the fully dense specification is computationally inefficient and susceptible to overfitting, especially when each $f_{ij}$ is independently parameterized. To better utilize the cross-variable structure, recent work adopts Channel Clustering (CC), which is equivalent to a block-diagonal constraint to $\rmF$. Essentially, the $D\times D$ entries of $\rmF$ are estimated by a significantly smaller number of modules, reflecting the low intrinsic dimensionality of many long-horizon high-dimensional time series. Through different grouping and coupling schemes for $\left\{f_{ij}\right\}$, one can instantiate different structural assumptions on inter-series dependence.

\subsection{AR-CR Structural Decoupling}
Modeling cross-dimension dependencies is challenging in multivariate forecasting. Autoregression (AR) is typically stable and persistent, whereas cross-relations (CR) are often regime-dependent and instantaneous. However, most dependency modeling methods either discard CR completely or homogeneously model the two.
Empirically (Fig.~\ref{fig:grad-variance-7}), joint optimization yields persistently higher gradient variance for CR, while AR gradients are smaller and mostly monotonically decaying. The high-variance CR updates inject non-negligible noise into the whole model, indicating different step-size/curvature needs; a shared optimizer entangles updates and lets CR noise bleed into AR, hindering convergence. These observations motivate a decoupled parameterization and training schedule.

\begin{figure*}[t]
  \centering
  \begin{subfigure}[t]{0.66\linewidth}
    \centering
    \includegraphics[width=\linewidth]{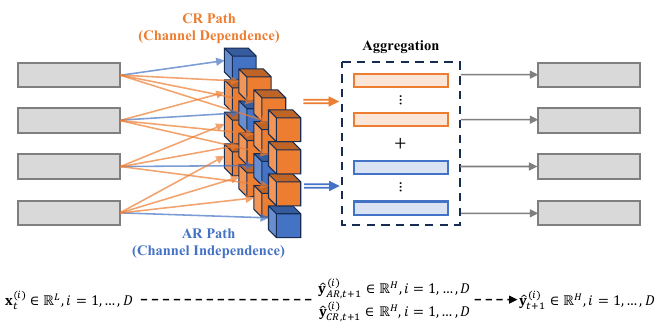}
    \caption{\framework\ Model Overview}
    \label{fig:arcr-a}
  \end{subfigure}\hfill
  \begin{subfigure}[t]{0.33\linewidth}
    \centering
    % 关键：只把图像上移，不影响子图 caption
    \raisebox{1.0cm}{\includegraphics[width=\linewidth]{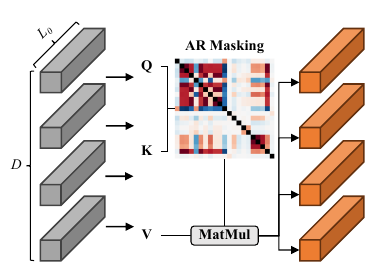}}
    \caption{Cross-Relation Self-Attention}
    \label{fig:arcr-b}
  \end{subfigure}
  \caption{Architecture of \framework. (a) Multivariate time series is passed into two parallel paths, the channel-independent AR path and the channel-dependent CR path. Outputs are summed to obtain the final prediction. (b) The cross-relation self-attention forms queries/keys/values from per-variable embeddings and an AR mask is applied to the attention matrix to suppress intra-series links.}
  \label{fig:arcr}
\end{figure*}

We therefore propose a dual-path design that explicitly separates AR and CR modeling, as illustrated in Figure \ref{fig:arcr}. Rather than estimating the full operator at once, we break down $\rmF$ into diagonal and off-diagonal components, $\rmF=\rmF_{\text{AR}}+\rmF_{\text{CR}}$, due to their distinct properties. $\rmF_{\text{AR}}=\mathrm{diag}(f_{11}, \dots, f_{DD})$ captures per-series AR patterns, and $\rmF_{\text{CR}}=\rmF - \rmF_{\text{AR}}$ encodes all cross-dimension dependencies.

\paragraph{Auto-Regression Path.}
For each variable $i$, we first apply Reversible Instance Normalization, $\mathrm{RevIN}$ \citep{kim2021reversible}, and fit a linear predictor $\hat{f}_{ii}$
\begin{equation}
    \hat{\rvy}_i^{\text{AR}} = \hat{f}_{ii}(\rvx_i)
\end{equation}
Parameters are not shared across $i$, enforcing channel independence. The AR path coincides with the RLinear model \citep{rlinear}. This simple model provides a clean separation of autoregression from cross-dimension effects. The AR path is trained with L1 regularization to prevent overfitting in long-horizon modeling.

\paragraph{Cross-Relation Path.}
We apply an inverted Transformer encoder that attends across variables. To prevent leakage of AR information into the CR module, we mask intra-series links in the attention matrix. Each individually normalized time series is compressed into a $L_0$-dimensional token via a linear projection,
\begin{equation}
    \rmZ_0 = \mathrm{Embedding}(\rmX_t)
\end{equation}
where $\rmZ_0\in \R^{D \times L_0}$ denotes the temporal embeddings of variables. $\rmZ_0$ is passed into the Multi-Head Self-Attention (MHSA) layer \citep{AttentionIsAllYouNeed}. For each head, the Cross-Relation Self-Attention (CRSA) is calculated as
\begin{equation}
    \mathrm{CRSA}(\rmZ_0)=\mathrm{softmax}(\frac{\rmQ\rmK^\top}{\sqrt{d_h}}+\rmM)\rmV
\end{equation}
where $\rmQ, \rmK, \rmV \in \R^{D \times d_h}$ denote queries, keys, and values, respectively. The attention learns {\em cross-variable} rather than cross-time dependencies. As such, the attention matrix $\rmA=(\rmQ\rmK^\top/\sqrt{d_h})$ represents the multivariate correlations, same as iTransformer \citep{liu2023itransformer}. The additional mask $\rmM=\mathrm{diag}(-\infty,\dots,-\infty)$ zeros out diagonal attention weights in $\mathrm{CRSA}$. This prevents the CR module from duplicating the AR function, thus enforcing a CR-only modeling. The remainder of the encoder follows the standard Transformer block \citet{AttentionIsAllYouNeed}
\begin{equation}
\begin{aligned}
    \rmZ_1 = \mathrm{LayerNorm}(\rmZ_0 + \mathrm{MHSA}(\rmZ_0)) \\
    \rmZ_2 = \mathrm{LayerNorm}(\rmZ_1 + \mathrm{MLP}(\rmZ_1))
\end{aligned}
\end{equation}
where $\mathrm{MLP}$ denotes a two-layer feedforward network. A channel-independent linear head maps the encoder output to the CR components $\hat{\rvy}_i^{\text{CR}}$ in the target sequence.

The final output is the denormalized sum of the AR and CR outputs. Our implementation uses a parameter-free $\mathrm{RevIN}$ layer, so the two paths share no trainable parameters.

% Our framework is different from the widely studied two-stage temporal-spatial modeling...

\subsection{Block Alternating Optimization}

\paragraph{Gradient Entanglement of Joint AR-CR Training.}
For simplicity, the following discussion considers the MSE loss function under minibatch  optimization, so gradients are stochastic. With perfect knowledge of cross-variable contributions $f_{ij},j=1,\dots,D$ for each variable $i$, the gradient for $\hat{f}_{ij}$ can be estimated as
\begin{equation}\label{gradtrue}
    \nabla_{\theta_{ij}}\mathcal{L}^*=-J_{ij}^\top \rvr_{ij},
\end{equation}
where $\mathcal{L}^*:=\frac{1}{2}\sum_{i,j=1}^D\Vert \rvr_{ij}\Vert^2_2$. $\theta_{ij}$ is the set of parameters for $\hat{f}_{ij}$, $J_{ij}$ is the Jacobian, and $\rvr_{ij}:=f_{ij}(\rvx_j) - \hat{f}_{ij}(\rvx_j)$ is the residual of the fitted projection $\hat{f}_{ij}$. Because (\ref{gradtrue}) is additively separable across $i,j$, the update is unaffected by other projections. The learned $\hat{f}_{ij}$ is therefore consistent with the true projection $f_{ij}$.

However, in real-world time series forecasting, such decomposition of residuals is infeasible. Without access to the true transition matrix $\rmF$, we only observe the aggregate residuals $\rvr_{i}:= \rvy_i - \hat{\rvy}_i = \sum_{j=1}^D \rvr_{ij}, i=1,\dots,D$. The estimated gradient mixes estimation errors from multiple projections,
\begin{equation}\label{gradmix}
    \nabla_{\theta_{ij}}\mathcal{L}=-J_{ij}^\top \rvr_i,
\end{equation}
where $\mathcal{L}:=\frac{1}{2}\sum_{i=1}^D\Vert \rvr_{i}\Vert^2_2$. Minimizing the sum of $\Vert \rvr_{i}\Vert^2_2$ does not guarantee the consistency between each individual $\hat{f}_{ij}$ and $f_{ij}$, but only the combined mapping over $j$. Equation \ref{gradmix} induces gradient entanglement between parameter blocks $\theta_{ii}$ and $\theta_{ij}$ through the shared residual $\rvr_i$ throughout training. 
% When series are relatively homogeneous and exhibit spurious correlations, this entanglement blurs the distinction between AR and CR, often leading to over-reliance on cross-variable information.
When series are relatively homogeneous and exhibit spurious correlations, joint training can couple the AR and CR updates, blurring the intended separation between temporal autoregression and cross-variable modeling (as shown in Figure~\ref{fig:grad-variance-7} and Appendix~\ref{app:extended-gv}), and often biasing optimization toward cross-variable signals when correlations are strong but weakly informative.

% When the model is unable to distinguish between $f_{ij}(\rvx_j)$ and $f_{ik}(\rvx_k)$ in $\rvy_i$, this entanglement of gradients can increase the difficulty of the learning process. 
% Inspired by the success of the CI assumption and the challenges encountered in the CD setting, it is reasonable to assume heterogeneity between AR and CR, as the latter is often estimated with large errors and instability.

% With the following assumptions, we examine the bias and variance introduced to the AR path by CR estimation errors, thereby revealing the benefit of alternating optimization for our dual-path structure.
% \begin{assumption}\label{assumption:norm}
% Let $\rvr_{-ii}:=\sum_{j\ne i}^D \rvr_{ij}$ be the sum of CR residuals with respect to variable $i$. Assume CR has significantly larger estimation errors conditional on $\rvx_i$
% \begin{equation}
%     \Vert \rvr_{ii}\Vert_2 < \E(\Vert \rvr_{-ii} \Vert_2 \mid \rvx_i).
% \end{equation}
% Note that $\E(\Vert \rvr_{ii}\Vert_2 \mid \rvx_i) = \Vert \rvr_{ii}\Vert_2$.
% \end{assumption}
Let $\rvr_{-ii}:=\sum_{j\ne i}^D \rvr_{ij}$ be the sum of CR residuals for variable $i$. For each AR projection $\hat{f}_{ii}$, the unbiasedness of (\ref{gradmix}) requires the conditional expectation of the bias term to be zero.
\begin{equation}\label{E_bias}
    \E(-J_{ii}^\top \rvr_{-ii}\mid\rvx_i) = -J_{ii}^\top \E(\rvr_{-ii}\mid\rvx_i) = 0,
\end{equation}
which requires $\E(\rvr_{-ii}\mid\rvx_i)=0$. However, due to the existence of cross-variable dependencies and the shared residual $\rvr_i$ among $\hat{f}_{ij}$, this requirement fails to hold unless projections $\hat{f}_{ij},j\ne i$ are constant with respect to $\rvx_i$ as in the CI setting. In the general CD setting, where $\E(\rvx_j\mid\rvx_i) \ne 0$ for some $j$, $\E(\rvr_{-ii}\mid\rvx_i)$ is a non-trivial function of $\rvx_i$.

% The bias in the mixed gradient may not be an issue for forecasting since we may still obtain unbiased predictions. Nonetheless, it can undermine the stability of the learning process. Under the assumption \ref{assumption:norm}, the expected relative error of the gradient has a lower bound
% \begin{equation}
%     \frac{\E(\Vert J_{ii}^\top \rvr_{-ii} \Vert_2 \mid \rvx_i)}{\E(\Vert J_{ii}^\top \rvr_{ii} \Vert_2 \mid \rvx_i)} \geq \frac{\sigma_{\min}(J_{ii})}{\sigma_{\max}(J_{ii})}\cdot \frac{\E(\Vert \rvr_{-ii} \Vert_2 \mid \rvx_i)}{\Vert \rvr_{ii} \Vert_2}.
% \end{equation}
% Normally, the smallest singular value $\sigma_{\min}(J_{ii})$ is non-zero. Therefore, the relative error induced by CR is driven by $\E(\Vert \rvr_{-ii} \Vert_2 \mid \rvx_i)$.

The bias in the mixed gradient may not pose a problem for forecasting since unbiased predictions can still be attainable. Nonetheless, it can degrade training stability. Conditional on $\rvx_i$, the covariance of (\ref{gradtrue}) is 0, whereas the covariance of the mixed gradient (\ref{gradmix}), $\Cov(-J_{ii}^\top \rvr_i \mid \rvx_i) = \Cov(-J_{ii}^\top \rvr_{-ii} \mid \rvx_i) \succeq 0$, is generally non-trivial. This additional covariance introduced by the CR contamination translates into noisier updates for the AR block. We consider the following assumption to investigate the unconditional covariance.
\begin{assumption}\label{assumption:cov}
Let $\Sigma_{ii}:=\Cov(-J_{ii}^\top \rvr_{ii})$, $\Sigma_{-ii}:=\Cov(-J_{ii}^\top \rvr_{-ii})$ be the covariance matrices of the true gradient and the CR contamination respectively. Assume the covariances satisfy
\begin{equation}\label{cov_trace}
    \Vert\Sigma_{-ii}\Vert_1 > 4 \Vert\Sigma_{ii}\Vert_1,
\end{equation}
where $\Vert\cdot\Vert_1$ is the trace norm.
\end{assumption}
The trace norm of gradient covariance serves as a metric for update turbulence, with the same definition as in \citet{NIPS2013SVRG}; \citet{CVPR2022VOG}. Equation \ref{cov_trace} suggests a lower bound for $\Vert\Sigma_{-ii}\Vert_1$ when the CR gradient variance is sufficiently large to affect the optimization of the AR path under various settings.

When a subset of variables in the input sequence is highly correlated, the indistinguishability arises and can lead to \eqref{cov_trace}. For instance, in a bivariate case with an instantaneous relationship $\rvx_i=\alpha\rvx_j + \epsilon,\ \forall t$ for some constant $\alpha$ while cross-dimension contributions are zero (i.e., $f_{ij}=f_{ji}=0$), $\Vert\Sigma_{-ii}\Vert_1$ can be inflated by misspecified $\hat{f}_{ij}$. Even without extreme dependence, small cross-dimension errors that arise intermittently can accumulate and materially raise gradient variance over training. To assess its impact on the AR gradient stability, consider the conditional covariance matrix of (\ref{gradmix})
\begin{equation}\label{cov_decomposition}
    \Cov(J_{ii}^\top \rvr_i) = \Sigma_{ii} + \Sigma_{-ii} + 2\Cov(J_{ii}^\top \rvr_{ii}, J_{ii}^\top \rvr_{-ii}).
\end{equation}
The trace norm of \eqref{cov_decomposition} satisfies
\begin{align*}
    \Vert\Cov(J_{ii}^\top \rvr_i)\Vert_1 & = \Vert\Sigma_{ii}\Vert_1 + \Vert\Sigma_{-ii}\Vert_1 \\&\qquad + 2\Tr(\Cov(J_{ii}^\top \rvr_{ii}, J_{ii}^\top \rvr_{-ii})) \\
    & \geq \Vert\Sigma_{ii}\Vert_1 + \Vert\Sigma_{-ii}\Vert_1 - 2\sqrt{\Vert\Sigma_{ii}\Vert_1\Vert\Sigma_{-ii}\Vert_1} \\
    & > \Vert\Sigma_{ii}\Vert_1.
\end{align*}
The first inequality uses the Cauchy-Schwarz inequality for covariance, with equality if and only if the true AR gradient $-J_{ii}^\top \rvr_{ii}$ and the CR contamination $-J_{ii}^\top \rvr_{-ii}$ are perfectly negatively correlated. The second inequality directly follows from assumption \ref{assumption:cov}. Hence, the mixed AR gradient is strictly less stable than the true AR gradient.

\paragraph{Alternating Training Strategy.}
To mitigate gradient entanglement, we separately optimize the AR and CR paths. Specifically, we repeat a two-step alternating optimization (AO) cycle until convergence.
\begin{itemize}
    \item Step 1. $\theta_{\text{AR}}^{(n+1)} \gets \argmin_{\theta_{\text{AR}}} \Vert \rmY_{t+1} - (\rmF_{\text{AR}} + \rmF_{\text{CR}}^{(n)}) * \rmX_t \Vert_2^2 + R_{\text{AR}}(\theta_{AR}).$
    \item Step 2. $\theta_{\text{CR}}^{(n+1)} \gets \argmin_{\theta_{\text{CR}}} \Vert \rmY_{t+1} - (\rmF_{\text{AR}}^{(n+1)} + \rmF_{\text{CR}}) * \rmX_t \Vert_2^2 + R_{\text{CR}}(\theta_{CR}).$
\end{itemize}
$R_{\text{AR}}(\cdot)$ and $R_{\text{CR}}(\cdot)$ are regularizers for the AR module and the CR module, respectively. In practice, we initialize two independent AMSGrad optimizers \citep{AMSGrad}. During training, we first activate the AR regularizer and optimize the AR parameters while keeping the CR parameters frozen; we then alternate the regularizer and freeze the AR parameters to optimize the CR parameters. Each subproblem is run for a small but non-trivial number of inner epochs on the entire training set before alternating. We use AMSGrad due to its convergence guarantee, aligning with alternating optimization algorithms that require sufficient descent and convergence of each subproblem \citep{BSUM,PALM}. For the same reason, we always update the AR path prior to the CR path, since the AR path uses a simpler model in our setting.

The proposed alternating training strategy isolates the two parameter blocks. Intuitively, each subproblem is easier to optimize than the original joint problem due to reduced dimensionality. It also enables distinct learning schedules tailored to AR and CR, which may intrinsically require different step sizes and regularization. The following proposition formalizes the stability benefit of alternating updates.

\begin{proposition}\label{prop}
    Suppose the loss function $\mathcal{L}(\theta_{\text{AR}}, \theta_{\text{CR}};B)=\ell(\theta_{\text{AR}}, \theta_{\text{CR}};B) + R_{\text{AR}}(\theta_{\text{AR}}) + R_{\text{CR}}(\theta_{\text{CR}})$ is the same under alternating and joint training, where $B$ denotes a random minibatch. Let $\Cov_{\text{alt}}(\nabla_{\theta_{ii}}\mathcal{L})$ be the covariance matrix of gradients for $\theta_{ii}$ under alternating training. Then $\Cov_{\text{alt}}(\nabla_{\theta_{ii}}\mathcal{L})\preceq \Cov(\nabla_{\theta_{ii}}\mathcal{L})$, where $\Cov(\nabla_{\theta_{ii}}\mathcal{L})$ is the joint training gradient covariance.
\end{proposition}
Note that since $\nabla_{\theta_{ii}}R_{\text{CR}}(\theta_{CR})=0$, using $\mathcal{L}$ is equivalent to applying the loss function in Step 1 when updating the AR path, and vice versa. By the law of total covariance, the gradient covariance under the joint training of $\theta_{ii}$ is
\begin{equation}
\begin{aligned}\label{law_total_cov}
    \Cov(\nabla_{\theta_{ii}}\mathcal{L}) = & \E_{\theta_{\text{CR}}}(\Cov(\nabla_{\theta_{ii}}\mathcal{L} \mid \theta_{\text{CR}})) + \\ 
    & \Cov_{\theta_{\text{CR}}}(\E(\nabla_{\theta_{ii}}\mathcal{L} \mid \theta_{\text{CR}})).
\end{aligned}
\end{equation}
In Step 1 of alternating training, $\theta_{\text{CR}}$ is held fixed, so the gradient covariance for $\theta_{ii}$ is $\Cov(\nabla_{\theta_{ii}}\mathcal{L} \mid \theta_{\text{CR}})$. Taking the expectation over the distribution of $\theta_{\text{CR}}$ yields
\begin{equation}\label{cov_alt}
    \Cov_{\text{alt}}(\nabla_{\theta_{ii}}\mathcal{L}) = \E_{\theta_{\text{CR}}}(\Cov(\nabla_{\theta_{ii}}\mathcal{L} \mid \theta_{\text{CR}})),
\end{equation}
which is precisely the first term in \eqref{law_total_cov}. The between-path source of gradient variability is removed. This particularly benefits the AR path, as the CR path usually has noticeably more parameters and slower convergence, making the second term in \eqref{law_total_cov} non-negligible.

% Algorithm
% \begin{algorithm}
%     \caption{AR-CR Alternating Optimization}
%     \label{algo:ao}
%     \begin{algorithmic}
%     \STATE \textbf{Initialize:} $\theta_{AR}^{(0)}$, $\theta_{CR}^{(0)}$
%     \FOR{$i=1$ \TO $n$}
%         \FOR{$i_1=1$ \TO $n_1$}
%             \STATE $\theta_{\text{AR}}^{(i_1)} \gets \argmin_{\theta_{\text{AR}}} \Vert \rmY_{t+1} - (\rmF_{\text{AR}} + \rmF_{\text{CR}}^{(i-1)}) * \rmX_t \Vert_2^2 $
%         \ENDFOR
%         \STATE $\theta_{\text{AR}}^{(i)} \gets \theta_{\text{AR}}^{(n_1)}$
%         \FOR{$i_2=1$ \TO $n_2$}
%             \STATE $\theta_{\text{CR}}^{(i_2)} \gets \argmin_{\theta_{\text{CR}}} \Vert \rmY_{t+1} - (\rmF_{\text{AR}}^{(i)} + \rmF_{\text{CR}}) * \rmX_t \Vert_2^2 $
%         \ENDFOR
%         \STATE $\theta_{\text{CR}}^{(i)} \gets \theta_{\text{CR}}^{(n_2)}$
%     \ENDFOR
%     \end{algorithmic}
% \end{algorithm}

% Two-branch instead of two-stage
% Compare with mixed and successive AR-CR

\section{Experiments}
% Full Results
\begin{table*}[t]

\caption{\textbf{Full Results.} Comparison across seven datasets (Weather, Traffic, Electricity, ETTh1, ETTh2, ETTm1, ETTm2) and four prediction horizons (96/192/336/720). Columns are ordered as \emph{Ours}, OLinear, TimeBase, iTransformer, RLinear, PatchTST, DLinear, and Informer. Metrics are MSE and MAE (lower is better). 
\textbf{Bold} numbers denote the best (lowest) performance in each row, while \underline{underlined} numbers denote the second best.}

\centering
\scriptsize
\begin{tabular}{c|c|cc|cc|cc|cc|cc|cc|cc|cc}
\hline
\multicolumn{2}{c|}{Model} & \multicolumn{2}{c|}{Ours} &
        \multicolumn{2}{c|}{OLinear} &
        \multicolumn{2}{c|}{TimeBase} &
        \multicolumn{2}{c|}{iTransformer} &
        \multicolumn{2}{c|}{RLinear} &
        \multicolumn{2}{c|}{PatchTST} &
        \multicolumn{2}{c|}{DLinear} &
        \multicolumn{2}{c}{Informer}
        \\ \hline

\multicolumn{2}{c|}{Metrics} & MSE & MAE & MSE & MAE & MSE & MAE & MSE & MAE & MSE & MAE & MSE & MAE & MSE & MAE & MSE & MAE \\ \hline

\multirow{4}{*}{\rotatebox{90}{Weather}} & 96 & \textbf{0.144} & \underline{0.195} & \underline{0.147} & \textbf{0.188} & 0.151 & 0.204 & 0.168 & 0.218 & 0.171 & 0.223 & 0.150 & 0.205 & 0.170 & 0.229 & 0.350 & 0.410 \\
 & 192 & \textbf{0.192} & \textbf{0.241} & \underline{0.193} & 0.244 & 0.195 & 0.248 & 0.210 & 0.255 & 0.216 & 0.260 & 0.194 & \underline{0.242} & 0.215 & 0.275 & 0.420 & 0.430 \\
 & 336 & \textbf{0.240} & \underline{0.281} & 0.247 & \textbf{0.276} & 0.244 & 0.282 & 0.260 & 0.291 & 0.261 & 0.294 & \underline{0.242} & 0.283 & 0.258 & 0.309 & 0.580 & 0.549\\
 & 720 & \textbf{0.310} & \textbf{0.329} & 0.320 & 0.333 & 0.317 & 0.336 & 0.331 & 0.341 & 0.323 & 0.339 & \underline{0.314} & \underline{0.332} & 0.319 & 0.359 & 0.920 & 0.699\\ \hline

\multirow{4}{*}{\rotatebox{90}{Traffic}} & 96 & \underline{0.355} & \underline{0.252} & \textbf{0.341} & \textbf{0.221} &  0.392 & 0.259 & 0.367 & 0.272 & 0.395 & 0.272 & 0.360 & \textbf{0.253} & 0.394 & 0.274 & 0.739 & 0.412 \\
 & 192 & \underline{0.378} & 0.265 & \textbf{0.368} & \textbf{0.233} & 0.413 & 0.274 & 0.382 & 0.269 & 0.407 & 0.276 & 0.379 & \underline{0.256} & 0.406 & 0.279 & 0.777 & 0.435\\
 & 336 & \textbf{0.395} & 0.275 & 0.396 & 0.242 & 0.427 & 0.287 & 0.395 & 0.279 & 0.416 & 0.282 & 0.397 & \underline{0.264} & 0.415 & 0.285 & 0.775 & 0.450\\
 & 720 & \textbf{0.429} & \underline{0.289} & 0.491 & \textbf{0.274} & 0.466 & 0.301 & 0.431 & 0.290 & 0.454 & 0.302 & 0.432 & 0.292 & 0.453 & 0.307 & 0.820 & 0.460\\ \hline

\multirow{4}{*}{\rotatebox{90}{Electricity}} & 96 & \underline{0.131} & 0.227 & 0.133 & \textbf{0.220} & 0.136 & 0.229 & 0.132 & 0.228 & 0.139 & 0.244 & \textbf{0.129} & \underline{0.222} & 0.135 & 0.232 & 0.300 & 0.399 \\
 & 192 & \underline{0.148} & \underline{0.242} & 0.154 & 0.244 & 0.159 & 0.255 & 0.153 & 0.248 & 0.150 & 0.243 & \textbf{0.147} & \textbf{0.240} & 0.149 & 0.245 & 0.327 & 0.418\\
 & 336 & \textbf{0.162} & \textbf{0.258} & 0.168 & 0.260 & 0.172 & 0.288 & 0.168 & 0.265 & 0.166 & 0.260 & \underline{0.163} & \underline{0.259} & 0.164 & 0.265 & 0.334 & 0.432\\
 & 720 & 0.204 & 0.292 & \textbf{0.194} & \textbf{0.283} & 0.219 & 0.301 & \underline{0.195} & \underline{0.286} & 0.212 & 0.300 & 0.197 & 0.290 & 0.198 & 0.295 & 0.356 & 0.429 \\ \hline

\multirow{4}{*}{\rotatebox{90}{ETTh1}} & 96 & \textbf{0.360} & \textbf{0.391} & \underline{0.366} & 0.400 & 0.384 & \underline{0.392} & 0.399 & 0.425 & 0.367 & 0.394 & 0.370 & 0.400 & 0.378 & 0.404 & 0.932 & 0.766\\
 & 192 & \textbf{0.397} & \textbf{0.415} & 0.407 & 0.427 & 0.432 & 0.462 & 0.426 & 0.442 & \underline{0.402} & 0.422 & 0.412 & 0.428 & 0.405 & \underline{0.417} & 1.001 & 0.780\\
 & 336 & \textbf{0.411} & \textbf{0.429} & 0.432 & 0.444 & 0.445 & 0.473 & 0.457 & 0.464 & \underline{0.419} & 0.617 & 0.421 & \underline{0.439} & 0.452 & 0.457 & 1.030 & 0.780\\
 & 720 & \textbf{0.435} & \textbf{0.453} & 0.452 & 0.470 & 0.449 & 0.482 & 0.630 & 0.574 & 0.451 & \underline{0.463} & \underline{0.447} & 0.467 & 0.502 & 0.513 & 1.139 & 0.852\\ \hline

\multirow{4}{*}{\rotatebox{90}{ETTh2}} & 96 & \textbf{0.273} & \underline{0.340} & 0.277 & \textbf{0.337} & 0.401 & 0.439 & 0.297 & 0.356 & \underline{0.275} & 0.341 & 0.279 & 0.341 & 0.282 & 0.348 & 1.542 & 0.957\\
 & 192 & \textbf{0.330} & \underline{0.381} & 0.351 & 0.383 & 0.451 & 0.467 & 0.377 & 0.405 & \underline{0.336} & \textbf{0.371} & 0.341 & 0.383 & 0.359 & 0.403 & 3.791 & 1.522\\
 & 336 & 0.333 & 0.392 & 0.370 & 0.402 & 0.458 & 0.482 & 0.424 & 0.440 & \textbf{0.324} & \underline{0.385} & \underline{0.328} & \textbf{0.383} & 0.440 & 0.454 & 4.200 & 1.640\\
 & 720 & \textbf{0.375} & \textbf{0.419} & 0.408 & 0.437 & 0.502 & 0.510 & 0.438 & 0.461 & 0.415 & 0.445 & \underline{0.379} & \underline{0.422} & 0.608 & 0.560 & 3.660 & 1.620\\ \hline

\multirow{4}{*}{\rotatebox{90}{ETTm1}} & 96 & \underline{0.290} & \textbf{0.337} & \textbf{0.279} & \underline{0.338} & 0.314 & 0.356 & 0.311 & 0.365 & 0.310 & 0.351 & 0.294 & 0.346 & 0.304 & 0.348 & 0.626 & 0.549 \\
 & 192 & \textbf{0.332} & \textbf{0.360} & 0.334 & \underline{0.362} & 0.339 & 0.370 & 0.348 & 0.385 & 0.338 & 0.367 & \underline{0.333} & 0.370 & 0.336 & 0.367 & 0.725 & 0.621\\
 & 336 & \textbf{0.365} & \textbf{0.380} & \underline{0.367} & 0.387 & 0.374 & 0.392 & 0.379 & 0.405 & 0.369 & \underline{0.385} & 0.370 & 0.392 & 0.374 & 0.395 & 1.002 & 0.745\\
 & 720 & \underline{0.423} & \textbf{0.412} & \textbf{0.407} & 0.417 & 0.424 & 0.425 & 0.443 & 0.444 & 0.429 & \underline{0.415} & 0.425 & 0.419 & 0.427 & 0.425 & 1.139 & 0.841\\ \hline

\multirow{4}{*}{\rotatebox{90}{ETTm2}} & 96 & \textbf{0.160} & \textbf{0.249} & \underline{0.162} & \underline{0.250} & 0.167 & 0.257 & 0.178 & 0.272 & 0.163 & 0.251 & 0.166 & 0.256 & 0.165 & 0.256 & 0.352 & 0.467\\
 & 192 & \textbf{0.214} & \textbf{0.286} & 0.219 & \underline{0.287} & 0.221 & 0.293 & 0.241 & 0.315 & \underline{0.217} & 0.292 & 0.223 & 0.296 & 0.226 & 0.306 & 0.599 & 0.579\\
 & 336 & \textbf{0.270} & \textbf{0.324} & 0.272 & 0.328 & 0.273 & 0.327 & 0.290 & 0.344 & \underline{0.271} & \underline{0.326} & 0.273 & 0.329 & 0.274 & 0.335 & 1.277 & 0.882\\
 & 720 & \underline{0.357} & \underline{0.382} & \textbf{0.345} & \textbf{0.372} & 0.368 & 0.389 & 0.376 & 0.397 & 0.360 & 0.387 & 0.361 & 0.385 & 0.380 & 0.408 & 2.892 & 1.219\\ \hline

\hline
% \multicolumn{2}{c|}{Count} 
% & x & x & x & x & x & x & x & x & x & x & x & x & x & x & x & x \\

\multicolumn{2}{c|}{Count}
& \multicolumn{2}{c|}{\textcolor{red}{49}}
& \multicolumn{2}{c|}{25}
& \multicolumn{2}{c|}{1}
& \multicolumn{2}{c|}{4}
& \multicolumn{2}{c|}{13}
& \multicolumn{2}{c|}{19}
& \multicolumn{2}{c|}{1}
& \multicolumn{2}{c}{0} \\
\hline

\end{tabular}

\label{tab:full-results}
\end{table*}

% Ablation study
\begin{table*}[t]

\caption{
% Ablation study of our method with and without Alternating Optimization (AO) optimization.  
% Each cell shows MSE/MAE, and \textbf{bold} indicates the better (lower) value for each metric.
Ablation study of our framework trained with alternating optimization (w/ AO) and with joint optimization of AR and CR (w/o AO). Each cell shows MSE/MAE; bold indicates the better value per metric.
}

\centering
\begin{subtable}[t]{\textwidth}
\caption{ETTh1, ETTh2, ETTm1, ETTm2}
\centering
\small
\begin{tabular}{c|cc|cc|cc|cc}
\hline
Horizon & \multicolumn{2}{c|}{ETTh1} & \multicolumn{2}{c|}{ETTh2} &
          \multicolumn{2}{c|}{ETTm1} & \multicolumn{2}{c}{ETTm2}\\
\cline{2-9}
(steps) & w/ AO & w/o AO & w/ AO & w/o AO & w/ AO & w/o AO & w/ AO & w/o AO \\ \hline
96  & \textbf{0.360}/\textbf{0.391} & 0.437/0.453 & \textbf{0.273}/\textbf{0.341} & 0.321/0.381 & \textbf{0.290}/\textbf{0.338} & 0.316/0.364 & \textbf{0.160}/\textbf{0.249} & 0.175/0.263 \\
192 & \textbf{0.397}/\textbf{0.415} & 0.471/0.473 & \textbf{0.330}/\textbf{0.381} & 0.371/0.414 & \textbf{0.331}/\textbf{0.360} & 0.359/0.387 & \textbf{0.214}/\textbf{0.286} & 0.243/0.313 \\
336 & \textbf{0.411}/\textbf{0.429} & 0.472/0.479 & \textbf{0.333}/\textbf{0.392} & 0.374/0.426 & \textbf{0.365}/\textbf{0.380} & 0.392/0.406 & \textbf{0.270}/\textbf{0.324} & 0.289/0.341 \\
720 & \textbf{0.435}/\textbf{0.453} & 0.510/0.506 & \textbf{0.375}/\textbf{0.419} & 0.440/0.463 & \textbf{0.423}/\textbf{0.412} & 0.449/0.433 & \textbf{0.357}/\textbf{0.382} & 0.381/0.401 \\ \hline
\end{tabular}
\end{subtable}

\vspace{2mm} % 子表间距离可自行调整

\begin{subtable}[t]{\textwidth}
\caption{Electricity, Traffic, Weather}
\centering
\small
\begin{tabular}{c|cc|cc|cc}
\hline
Horizon & \multicolumn{2}{c|}{Electricity} & \multicolumn{2}{c|}{Traffic} & \multicolumn{2}{c}{Weather}\\
\cline{2-7}
(steps) & w/ AO & w/o AO & w/ AO & w/o AO & w/ AO & w/o AO \\ \hline
96  & \textbf{0.131}/\textbf{0.227} & 0.137/0.233 & \textbf{0.355}/\textbf{0.252} & 0.364/0.258 & \textbf{0.144}/\textbf{0.195} & 0.157/0.211 \\
192 & \textbf{0.148}/\textbf{0.242} & 0.161/0.256 & \textbf{0.380}/\textbf{0.265} & 0.386/0.269 & \textbf{0.192}/\textbf{0.241} & 0.203/0.253 \\
336 & \textbf{0.162}/\textbf{0.258} & 0.179/0.274 & \textbf{0.395}/0.275 & 0.396/\textbf{0.274} & \textbf{0.240}/\textbf{0.281} & 0.249/0.288 \\
720 & 0.204/0.292 & \textbf{0.198}/\textbf{0.291} & \textbf{0.429}/\textbf{0.292} & 0.481/0.301 & \textbf{0.310}/\textbf{0.329} & 0.321/0.339 \\ \hline
\end{tabular}
\end{subtable}

\label{tab:ao-ablation}
\end{table*}

\subsection{Experimental Overview}
We conduct experiments on standard long-term time series forecasting benchmarks to evaluate the effectiveness of decoupling autoregressive and cross-variable modeling via alternating optimization. Specifically, we examine whether alternating optimization improves forecasting accuracy over joint training, whether the gains are consistent across datasets and horizons, and how \framework\ compares with recent state-of-the-art linear and attention-based baselines. We further analyze optimization behavior using gradient statistics to elucidate the impact of alternating updates.

\subsection{Experimental Setup}
\paragraph{Datasets.} 
We evaluate on seven standard LTSF benchmarks: \textbf{Weather}, \textbf{Traffic}, \textbf{Electricity}, and the ETT family (\textbf{ETTh1}, \textbf{ETTh2}, \textbf{ETTm1}, \textbf{ETTm2}). 
Each dataset is chronologically split into train/validation/test sets, and we forecast horizons of $\{96, 192, 336, 720\}$ steps. 
The input length is fixed to 512, following the convention of PatchTST~\citep{Yuqietal-2023-PatchTST}.
This setting provides a sufficiently long receptive field to capture both autoregressive stability and cross-variable interactions. 
Detailed dataset descriptions and statistics are provided in Appendix~\ref{app:dataset}.

\paragraph{Baselines.} 
We benchmark against representative methods spanning different modeling paradigms: 
\emph{RLinear} \citep{rlinear} and \emph{DLinear} \citep{dlinear} (channel-independent/pooled linear models), 
\emph{PatchTST} \citep{Yuqietal-2023-PatchTST} (patch-based CI/CM Transformer), 
\emph{iTransformer} \citep{liu2023itransformer} (variable-token attention for cross-variable modeling), 
\emph{Informer} \citep{haoyietal-informer-2021} (sparse attention), 
\emph{TimeBase} \citep{huang2025timebase} (autoregression-centric backbone),
and OLinear~\citep{yue2025olinear}, a recent SOTA linear forecaster operating in an orthogonally transformed domain.
These baselines cover both autoregressive and cross-variable approaches, providing a comprehensive set of alternative state-of-the-art approaches to evaluate our framework. 

\paragraph{Implementation Details.} 
In our dual-path framework, we instantiate the \textbf{AR path with RLinear}~\citep{rlinear} and the \textbf{CR path with iTransformer}~\citep{liu2023itransformer}, aiming to highlight the effect of \emph{alternating optimization} rather than architectural refinements. 
This instantiation is chosen for clarity rather than necessity: RLinear and iTransformer form a clean decomposition between autoregressive and cross-variable modeling, which allows us to isolate the effect of alternating optimization. More generally, our framework is \textbf{architecture-agnostic}: both the AR and CR components can be replaced by alternative modules without modifying the optimization procedure.
RLinear cleanly models channel-wise autoregression, while iTransformer captures cross-variable dependencies via variable-level self-attention. 
Both paths use RevIN normalization and are additively combined after denormalization. 
We adapt Alternating Optimization (AO) in training: AR parameters are updated with CR frozen and vice versa on each mini-batch, each subproblem optimized by AMSGrad with early stopping. We fix the AO schedule to \textbf{10 AR updates} and \textbf{2 CR updates} per mini-batch, reflecting the greater stability of autoregressive dynamics and the higher variance of cross-variable modeling.
We scale $R_{\text{AR}}(\cdot)$ and $R_{\text{CR}}(\cdot)$ by prediction length to account for the sensitivity of the $\ell_1$ regularizer to parameter count.

\paragraph{Metrics.}
We report Mean Squared Error (MSE) and Mean Absolute Error (MAE) averaged across all variates, following LTSF conventions. Lower values indicate better predictive performance.

\subsection{Main Results}
Table~\ref{tab:full-results} summarizes results across all datasets and horizons.
\textbf{Overall, \framework\ attains SOTA or second-best performance on the vast majority of settings.}
Quantitatively, across all datasets, horizons, and both metrics (56 evaluation slots), \framework\ achieves the strongest overall performance with 49 top-two finishes (25 best + 24 second-best), substantially outperforming all baselines. The closest competitor, OLinear, attains 25 top-two results, followed by PatchTST with 19.
On the \emph{ETT family}, \framework\ is consistently strong—SOTA on ETTh1 (all horizons) and ETTm2 (all horizons), SOTA on ETTh2 at 96/192 and 720 (both metrics), and SOTA/runner-up on ETTm1 (SOTA at 96/192/336; best MAE and second-best MSE at 720).
On \emph{Weather}, \framework\ is SOTA at 96, 192, and 720 on both metrics (and best MSE at 336).
On \emph{Electricity}, it is competitive—SOTA at 336 and second-best at 96/192, while iTransformer leads at 720.
On \emph{Traffic}, \framework\ leads on MSE at 96 and is generally second-best at 192/336; at 720, iTransformer (MSE) and PatchTST (MAE) edge out \framework.
Taken together, these results show that decoupling AR and CR, paired with alternating optimization, reliably improves stability and accuracy, with the clearest gains on ETT and Weather, especially at longer horizons.

\subsection{Ablation Studies}

We conduct ablation studies to better understand the contributions of the two key components in our framework: the decoupled AR and CR paths, and the alternating optimization strategy. 
The first set of experiments disentangle the effect of each path by comparing against their standalone counterparts (RLinear and iTransformer), while the second set isolates the role of alternating optimization (AO) by contrasting alternating training with joint optimization. 
Together, these analyses clarify how architectural decoupling and optimization strategy jointly contribute to the performance of \framework. 

\paragraph{AR vs.\ CR Path Contributions.}
To more closely probe the dual-path design, we compare RLinear (AR-only) and iTransformer (CR-only) to \framework\ in Table~\ref{tab:full-results}.
RLinear excels on \emph{Electricity}, where strong periodicity makes per-series autoregression sufficient, but it lags on \emph{Weather} and \emph{Traffic}—datasets where inter-variable structure is essential.
Conversely, iTransformer benefits datasets with pronounced cross-variable coupling (e.g., \emph{Weather}, short/mid-horizon \emph{Traffic}) yet underperforms on the \emph{ETT} family, where stable autoregression dominates signal.
\framework\ reconciles these regimes: the AR path supplies low-variance, persistent dynamics while the CR path selectively captures intermittent cross-relations; alternating optimization coordinates the two so that CR updates do not contaminate AR gradients.
Empirically, \framework\ achieves best or second-best accuracy in the majority of datasets and horizons, with the largest margins on longer horizons in \emph{ETT} and \emph{Weather}, indicating that AR/CR complementarity—together with alternating updates—yields both greater stability and better forecasting efficacy.
These results indicate that alternating optimization is most beneficial when AR and CR signals coexist but compete unevenly, a setting where joint training tends to entangle their gradients and blur their inductive roles.

\paragraph{Effect of Alternating Optimization.}
Table~\ref{tab:ao-ablation} contrasts alternating optimization (AO) with joint training (single optimizer for AR and CR).
Overall, AO lowers both MSE and MAE across datasets and horizons, with the largest gains emerging at long horizons where joint training is most unstable (e.g., \emph{ETTh1} at 720: (0.435/0.453) vs.\ (0.510/0.506)).
Improvements are also consistent on \emph{ETTh2}, \emph{ETTm1}, and \emph{ETTm2}, and hold on \emph{Traffic} for short/mid horizons.
A few narrow exceptions remain (e.g., \emph{Electricity} at 720 and isolated MAE ties around \emph{Traffic}–336), indicating that while AO is broadly beneficial, extremely CI-friendly regimes can slightly favor joint tuning.
These results support our analysis: alternating updates mitigate AR–CR gradient entanglement, yielding more stable optimization and better long-horizon accuracy.

\section{Conclusions}
In this paper, we have introduced \framework, a dual-path framework for multivariate time series forecasting that \emph{decouples} autoregression (AR) and cross-relation (CR) modeling and coordinates them via \emph{alternating optimization}. 
Grounded in an analysis of gradient entanglement under joint AR–CR training, we instantiated the framework with RLinear for AR and modified iTransformer for CR to emphasize optimization over architectural novelty. 
Across seven multivariate LTSF benchmarks and four horizons, \framework\ delivers competitive or superior accuracy, with the largest gains at long horizons where joint training is most unstable. 
Ablations confirm that AR/CR separation is complementary and that alternating optimization is the key driver of stability and performance.
In summary, \framework\ is the first deep learning framework to explicitly decouple autoregression and cross-variable dependency via alternating optimization, supported by a theoretical analysis of gradient entanglement. 
Extensive experiments on seven benchmarks show consistent improvements over strong linear, Transformer-based, and hybrid baselines. 
Finally, we highlight training schedules as a design variable, suggesting a broader paradigm where optimization principles inform neural network architecture.

\bibliography{arxiv}
\bibliographystyle{icml2026}

%%%%%%%%%%%%%%%%%%%%%%%%%%%%%%%%%%%%%%%%%%%%%%%%%%%%%%%%%%%%%%%%%%%%%%%%%%%%%%%
%%%%%%%%%%%%%%%%%%%%%%%%%%%%%%%%%%%%%%%%%%%%%%%%%%%%%%%%%%%%%%%%%%%%%%%%%%%%%%%
% APPENDIX
%%%%%%%%%%%%%%%%%%%%%%%%%%%%%%%%%%%%%%%%%%%%%%%%%%%%%%%%%%%%%%%%%%%%%%%%%%%%%%%
%%%%%%%%%%%%%%%%%%%%%%%%%%%%%%%%%%%%%%%%%%%%%%%%%%%%%%%%%%%%%%%%%%%%%%%%%%%%%%%
\newpage
\appendix
\onecolumn

% \section{LLM Usage Disclosure}

% In accordance with the policy on large language models (LLMs), we disclose the use of ChatGPT as a writing assist tool. 
% Specifically, the LLM was employed to polish the presentation of text, improve clarity, and refine phrasing. 
% The authors carefully reviewed and edited all LLM-assisted text to ensure accuracy and alignment with the intended scientific contributions. 
% No part of the ideation, or substantive analysis relied on the LLM.

\section{Dataset Details}
\label{app:dataset}

\begin{table}[h]
\centering
\small
\begin{tabular}{lccccc}
\hline
Dataset & Variates & Freq. & Input Len. & Pred. Len. & Domain  \\
\hline
ETTh1   & 7   & Hourly   & 512 & 96–720  & Electricity \\
ETTh2   & 7   & Hourly   & 512 & 96–720  & Electricity \\
ETTm1   & 7   & 15 min   & 512 & 96–720  & Electricity \\
ETTm2   & 7   & 15 min   & 512 & 96–720  & Electricity \\
Weather & 21  & 10 min   & 512 & 96–720  & Weather     \\
Electricity & 321 & Hourly & 512 & 96–720 & Electricity \\
Traffic & 862 & Hourly   & 512 & 96–720  & Transport  \\
\hline
\end{tabular}
\caption{\textbf{Dataset statistics.}}
\label{tab:dataset-statistics}
\end{table}

We evaluate long-term forecasting performance on seven widely used multivariate benchmarks: \textbf{Weather}, \textbf{Traffic}, \textbf{Electricity}, and the ETT family (\textbf{ETTh1}, \textbf{ETTh2}, \textbf{ETTm1}, \textbf{ETTm2}). 
These datasets are standard in the LTSF community and cover diverse domains including weather monitoring, traffic flow, and energy consumption. 
Following the established protocol of \citet{Yuqietal-2023-PatchTST, wu2021autoformer}, we split the data chronologically into training/validation/test sets with a ratio of $6{:}2{:}2$ for the ETT datasets and $7{:}1{:}2$ for the others. 
The input length is fixed to 512 across all experiments, and prediction horizons are $\{96, 192, 336, 720\}$. 
Key statistics are summarized in Table~\ref{tab:dataset-statistics}.  

\paragraph{Weather.}  
Records 21 meteorological indicators (e.g., temperature, humidity, wind speed) every 10 minutes throughout 2020 in Germany.  

\paragraph{Traffic.}  
Hourly road occupancy rates measured by 862 sensors on San Francisco Bay Area freeways between 2015 and 2016.  

\paragraph{Electricity.}  
Hourly electricity consumption (kWh) of 321 customers from 2012 to 2014.  

\paragraph{ETT.}  
The Electricity Transformer Temperature datasets include two hourly datasets (ETTh1, ETTh2) and two 15-minute datasets (ETTm1, ETTm2). 
Each contains seven oil temperature and load features collected from electricity transformers between July 2016 and July 2018.

\newpage
\section{Extended Gradient-Variance Plots}
\label{app:extended-gv}

% -------------------------------------------------------
\subsection{Prediction length = 96}
\begin{figure}[H]
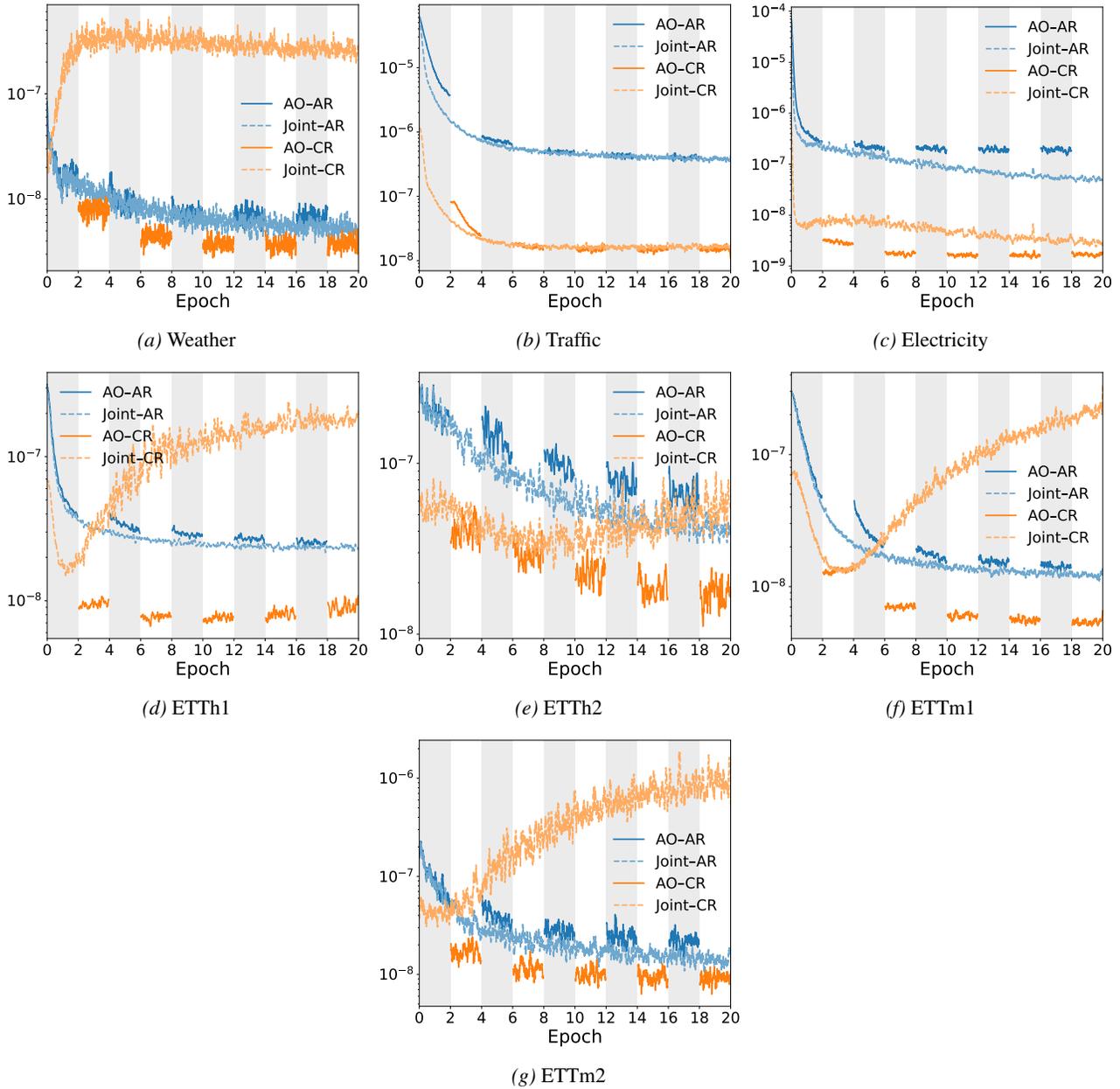

  \centering
  % Row 1
  \begin{subfigure}[t]{0.33\linewidth}
    \centering\includegraphics[width=\linewidth]{figures/gv_weather_512_96.pdf}
    \caption{Weather}
  \end{subfigure}%
  \begin{subfigure}[t]{0.33\linewidth}
    \centering\includegraphics[width=\linewidth]{figures/gv_traffic_512_96.pdf}
    \caption{Traffic}
  \end{subfigure}%
  \begin{subfigure}[t]{0.33\linewidth}
    \centering\includegraphics[width=\linewidth]{figures/gv_electricity_512_96.pdf}
    \caption{Electricity}
  \end{subfigure}

  \vspace{0.5em}

  % Row 2
  \begin{subfigure}[t]{0.33\linewidth}
    \centering\includegraphics[width=\linewidth]{figures/gv_etth1_512_96.pdf}
    \caption{ETTh1}
  \end{subfigure}%
  \begin{subfigure}[t]{0.33\linewidth}
    \centering\includegraphics[width=\linewidth]{figures/gv_etth2_512_96.pdf}
    \caption{ETTh2}
  \end{subfigure}%
  \begin{subfigure}[t]{0.33\linewidth}
    \centering\includegraphics[width=\linewidth]{figures/gv_ettm1_512_96.pdf}
    \caption{ETTm1}
  \end{subfigure}

  \vspace{0.5em}

  % Row 3 (single; centered with empty boxes)
  \begin{subfigure}[t]{0.33\linewidth}\centering\mbox{}\end{subfigure}%
  \begin{subfigure}[t]{0.33\linewidth}
    \centering\includegraphics[width=\linewidth]{figures/gv_ettm2_512_96.pdf}
    \caption{ETTm2}
  \end{subfigure}%
  \begin{subfigure}[t]{0.33\linewidth}\centering\mbox{}\end{subfigure}

  \caption{\textbf{Prediction length $=96$.} Variance of AR/CR gradients under joint training across seven datasets. Higher variance indicates greater training instability, motivating alternating optimization in \framework.}
  \label{fig:gv-7-96}
\end{figure}

% -------------------------------------------------------
\newpage
\subsection{Prediction length = 192}
\begin{figure}[H]
  \centering
  % Row 1
  \begin{subfigure}[t]{0.33\linewidth}
    \centering\includegraphics[width=\linewidth]{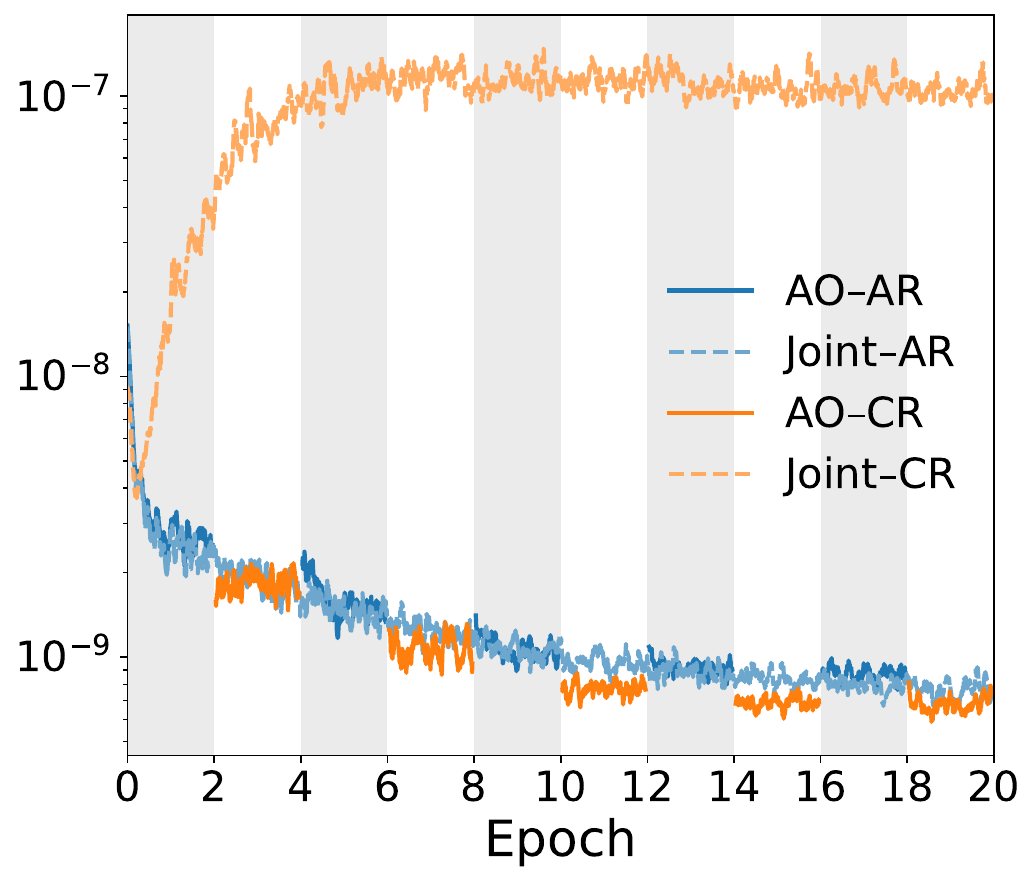}
    \caption{Weather}
  \end{subfigure}%
  \begin{subfigure}[t]{0.33\linewidth}
    \centering\includegraphics[width=\linewidth]{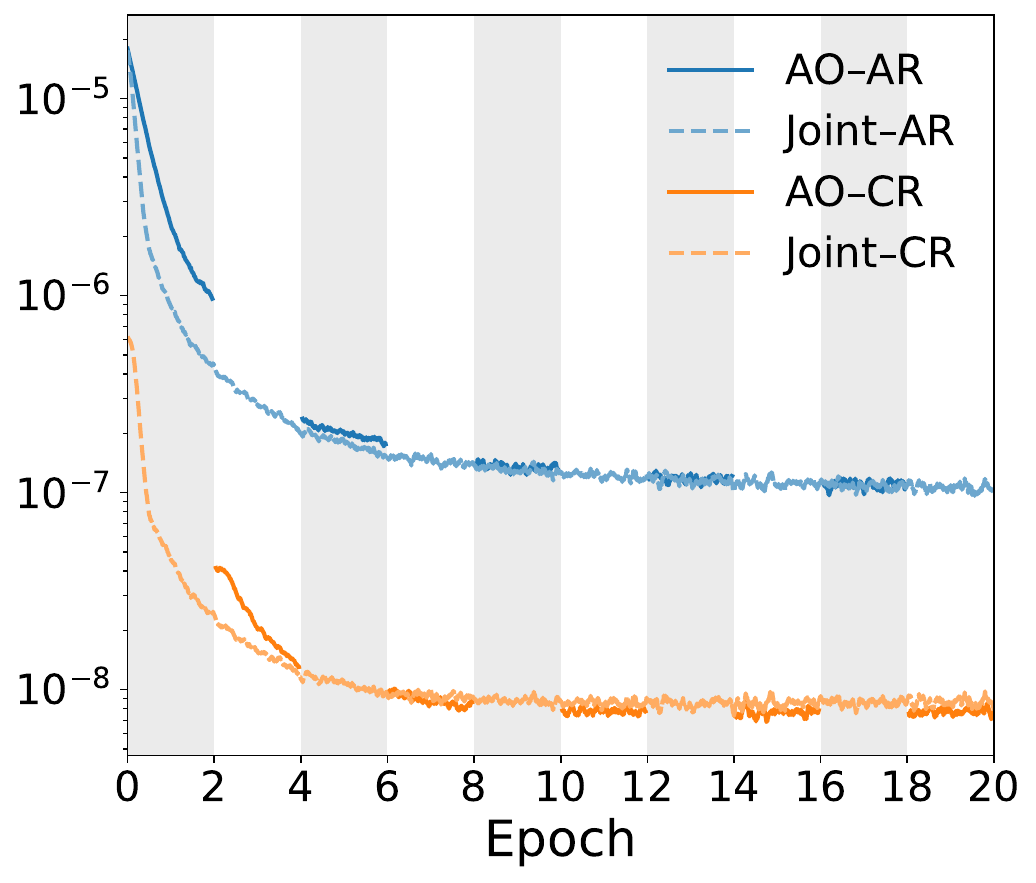}
    \caption{Traffic}
  \end{subfigure}%
  \begin{subfigure}[t]{0.33\linewidth}
    \centering\includegraphics[width=\linewidth]{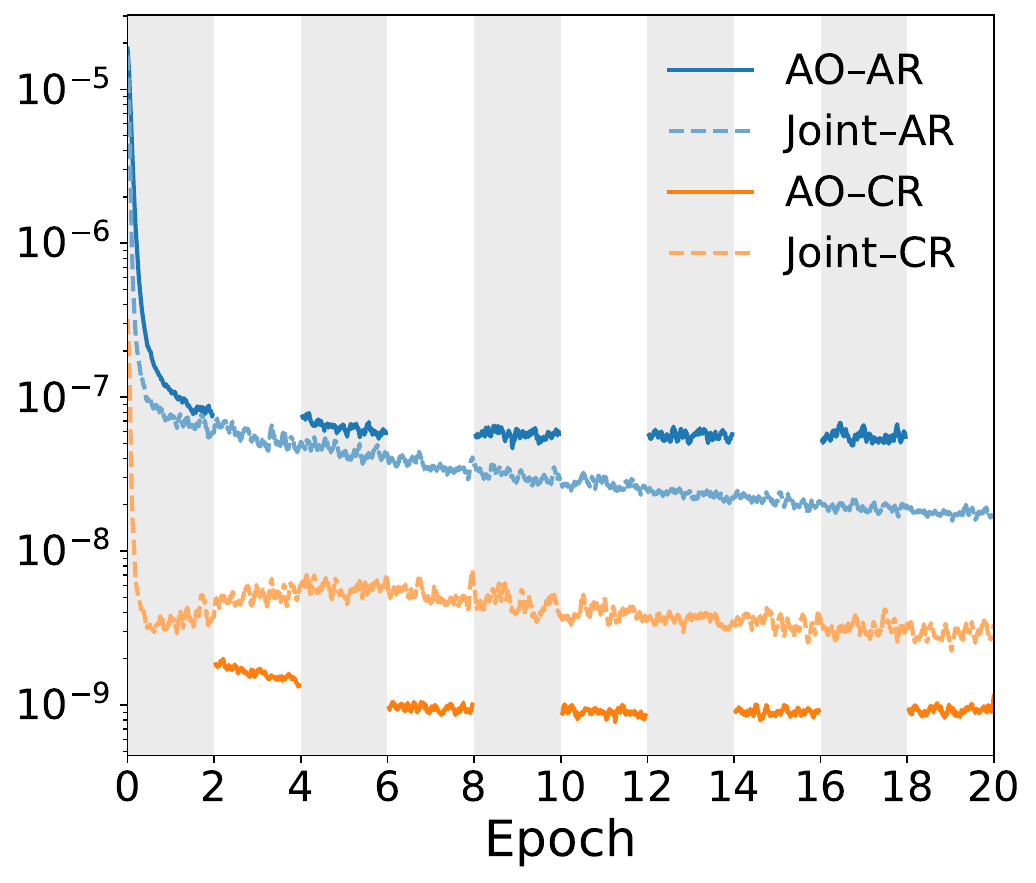}
    \caption{Electricity}
  \end{subfigure}

  \vspace{0.5em}

  % Row 2
  \begin{subfigure}[t]{0.33\linewidth}
    \centering\includegraphics[width=\linewidth]{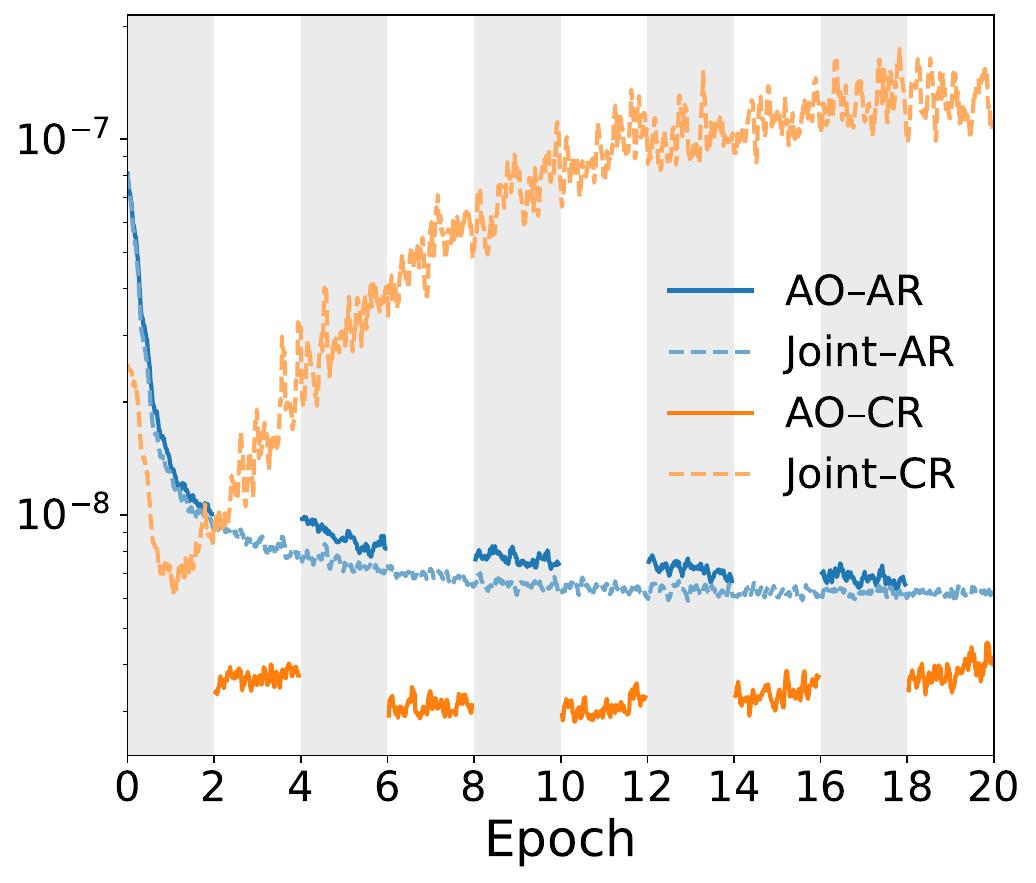}
    \caption{ETTh1}
  \end{subfigure}%
  \begin{subfigure}[t]{0.33\linewidth}
    \centering\includegraphics[width=\linewidth]{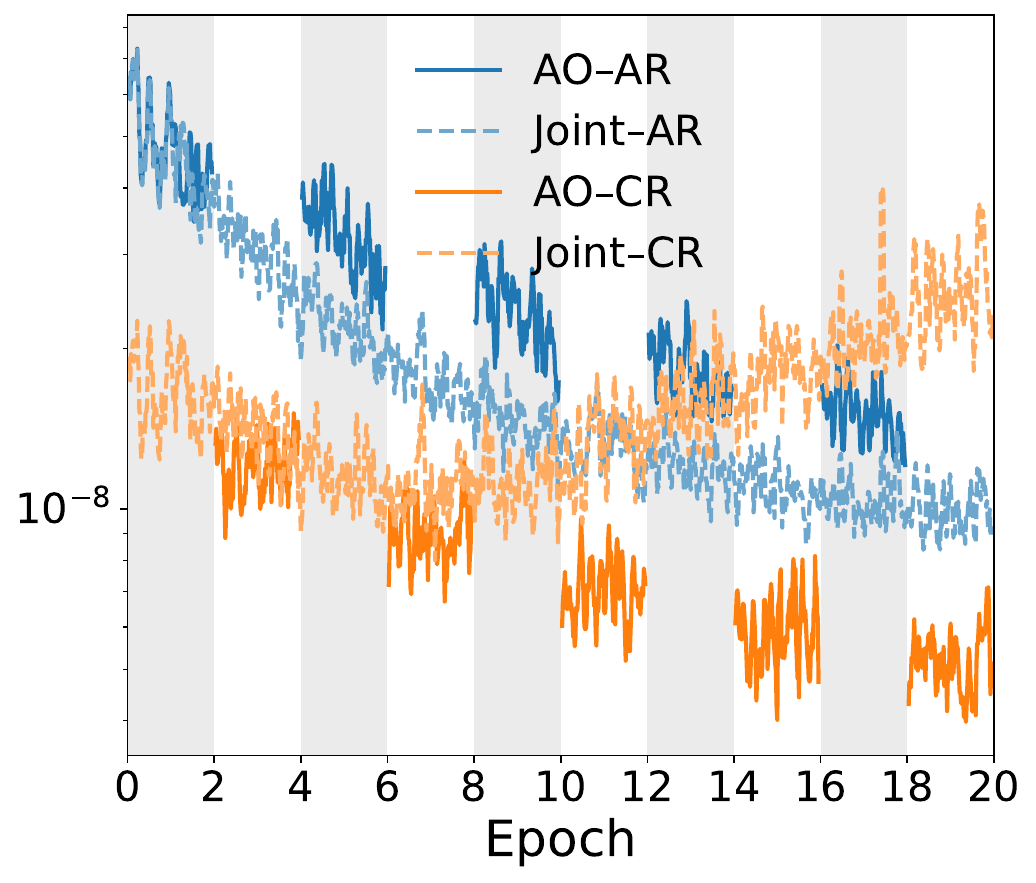}
    \caption{ETTh2}
  \end{subfigure}%
  \begin{subfigure}[t]{0.33\linewidth}
    \centering\includegraphics[width=\linewidth]{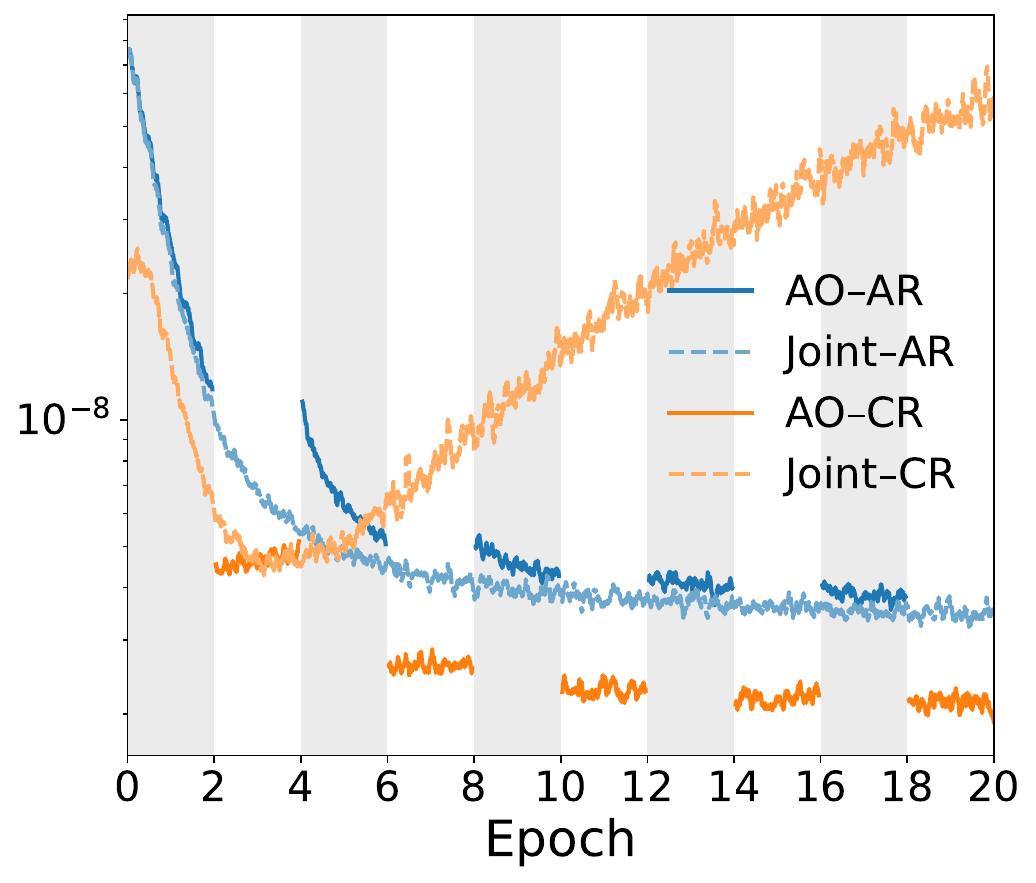}
    \caption{ETTm1}
  \end{subfigure}

  \vspace{0.5em}

  % Row 3 (single)
  \begin{subfigure}[t]{0.33\linewidth}\centering\mbox{}\end{subfigure}%
  \begin{subfigure}[t]{0.33\linewidth}
    \centering\includegraphics[width=\linewidth]{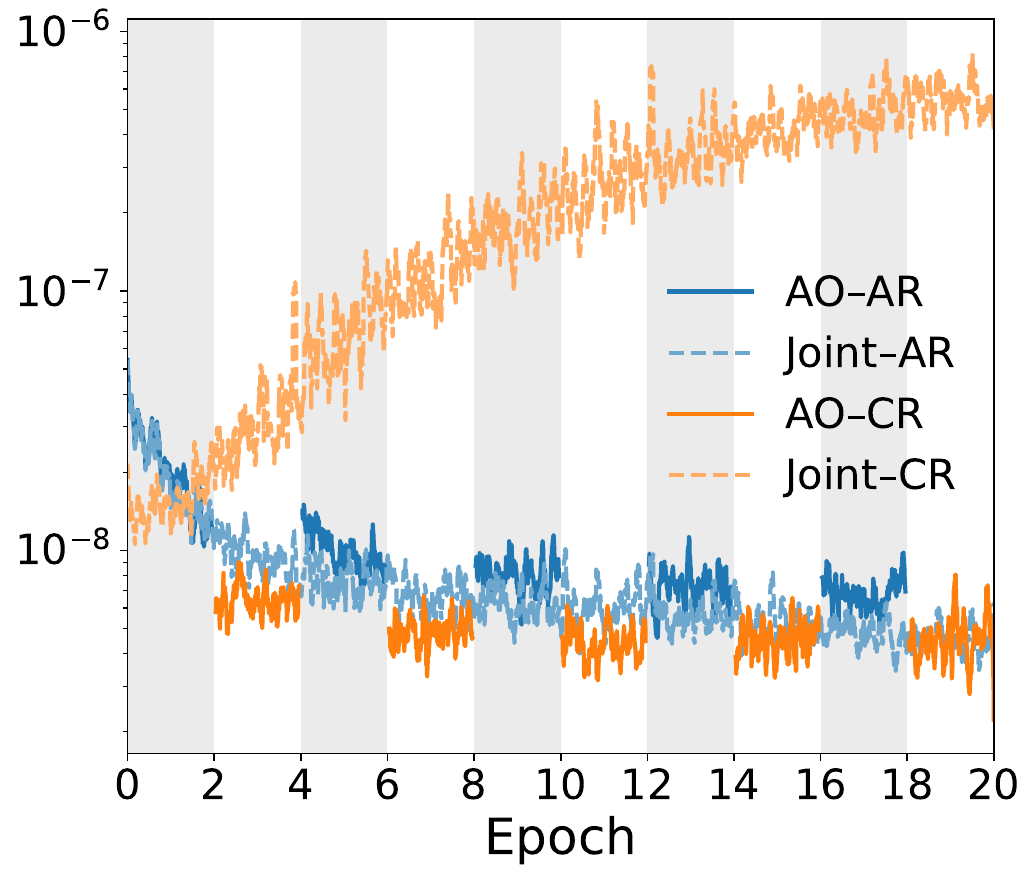}
    \caption{ETTm2}
  \end{subfigure}%
  \begin{subfigure}[t]{0.33\linewidth}\centering\mbox{}\end{subfigure}

  \caption{\textbf{Prediction length $=192$.} Variance of AR/CR gradients under joint training across seven datasets.}
  \label{fig:gv-7-192}
\end{figure}

% -------------------------------------------------------
\newpage
\subsection{Prediction length = 336}
\begin{figure}[H]
  \centering
  % Row 1
  \begin{subfigure}[t]{0.33\linewidth}
    \centering\includegraphics[width=\linewidth]{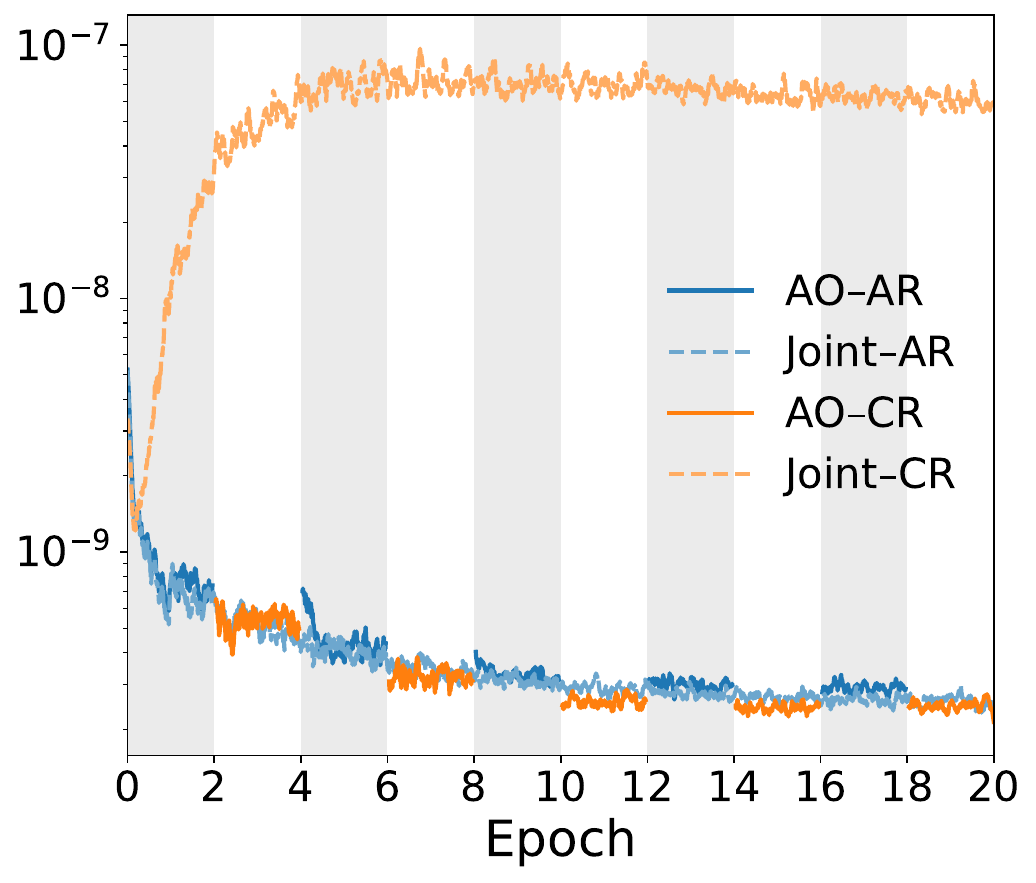}
    \caption{Weather}
  \end{subfigure}%
  \begin{subfigure}[t]{0.33\linewidth}
    \centering\includegraphics[width=\linewidth]{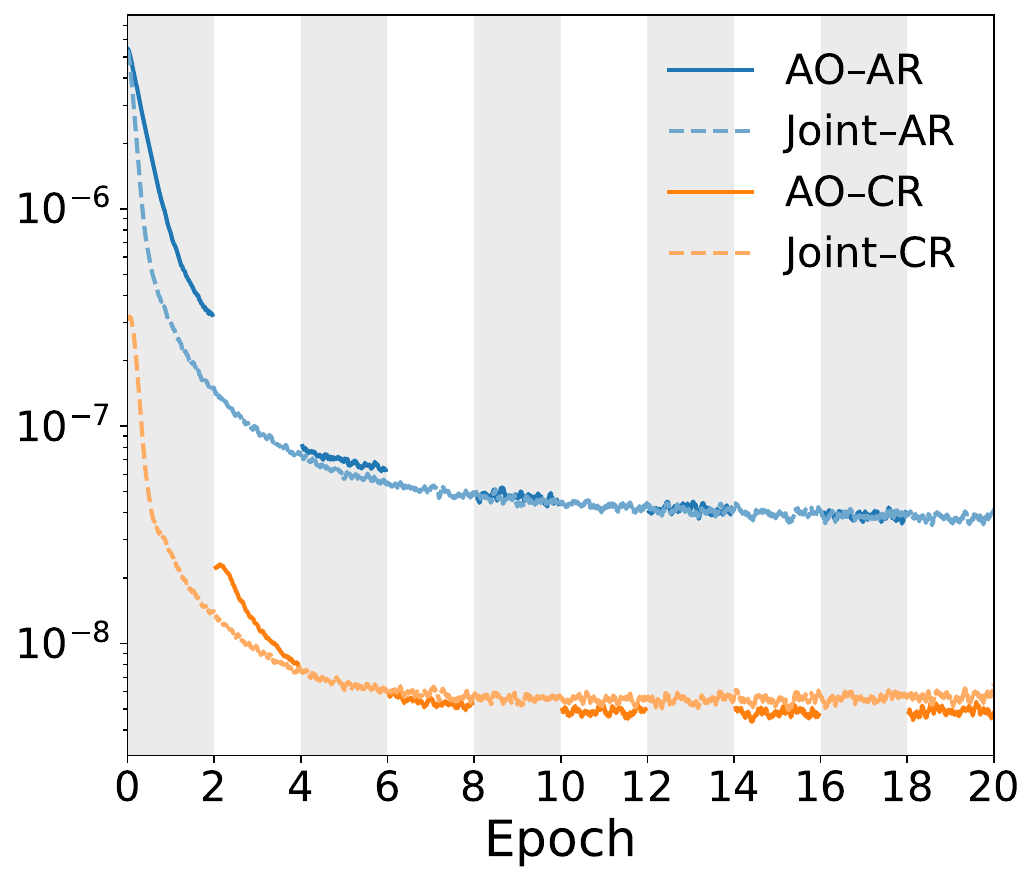}
    \caption{Traffic}
  \end{subfigure}%
  \begin{subfigure}[t]{0.33\linewidth}
    \centering\includegraphics[width=\linewidth]{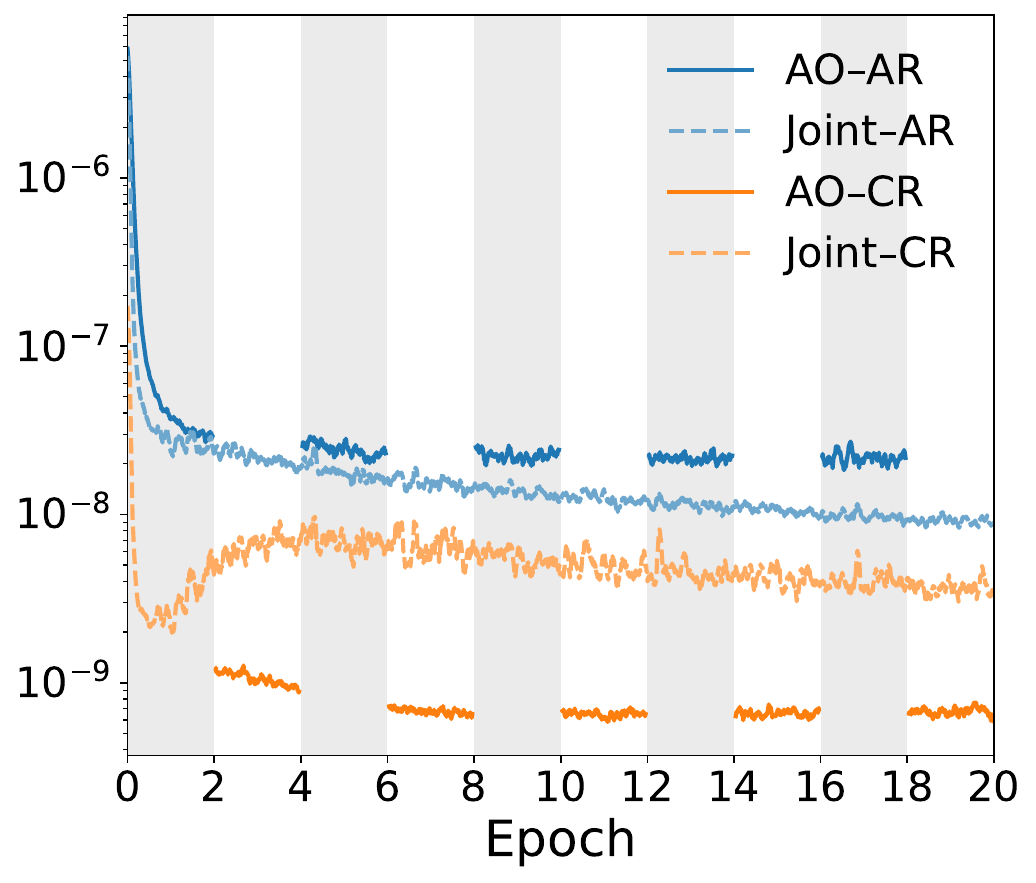}
    \caption{Electricity}
  \end{subfigure}

  \vspace{0.5em}

  % Row 2
  \begin{subfigure}[t]{0.33\linewidth}
    \centering\includegraphics[width=\linewidth]{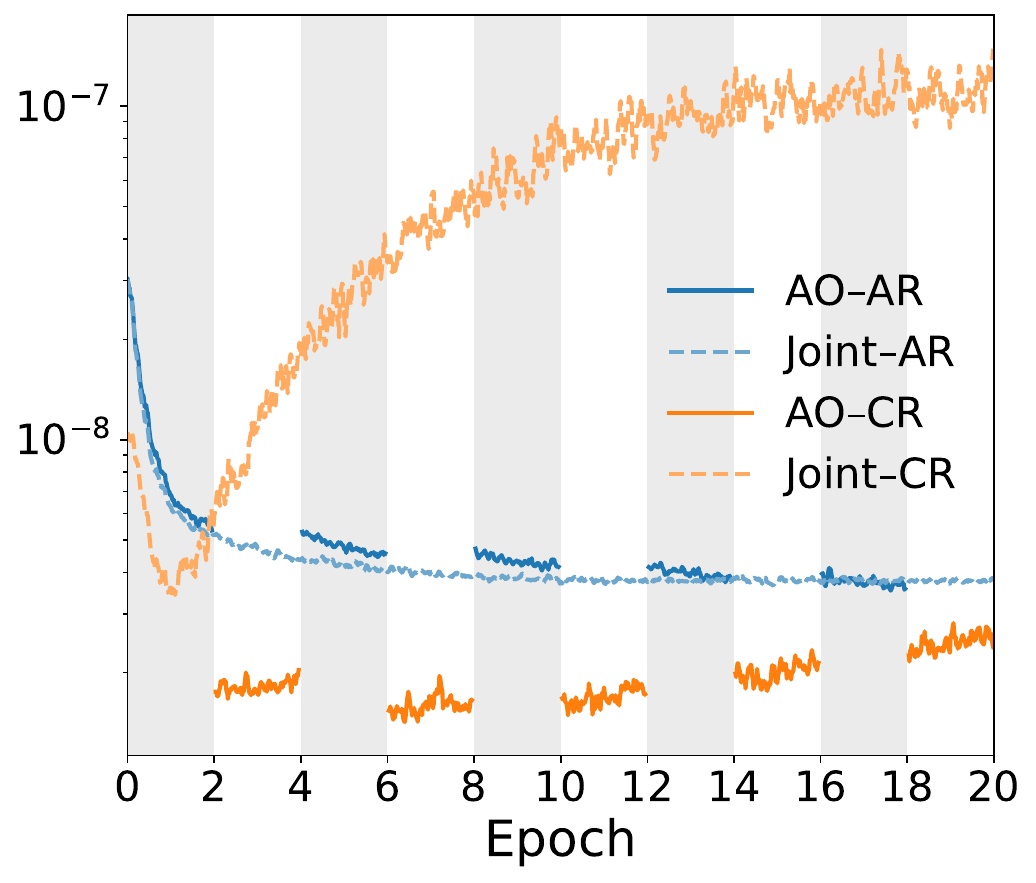}
    \caption{ETTh1}
  \end{subfigure}%
  \begin{subfigure}[t]{0.33\linewidth}
    \centering\includegraphics[width=\linewidth]{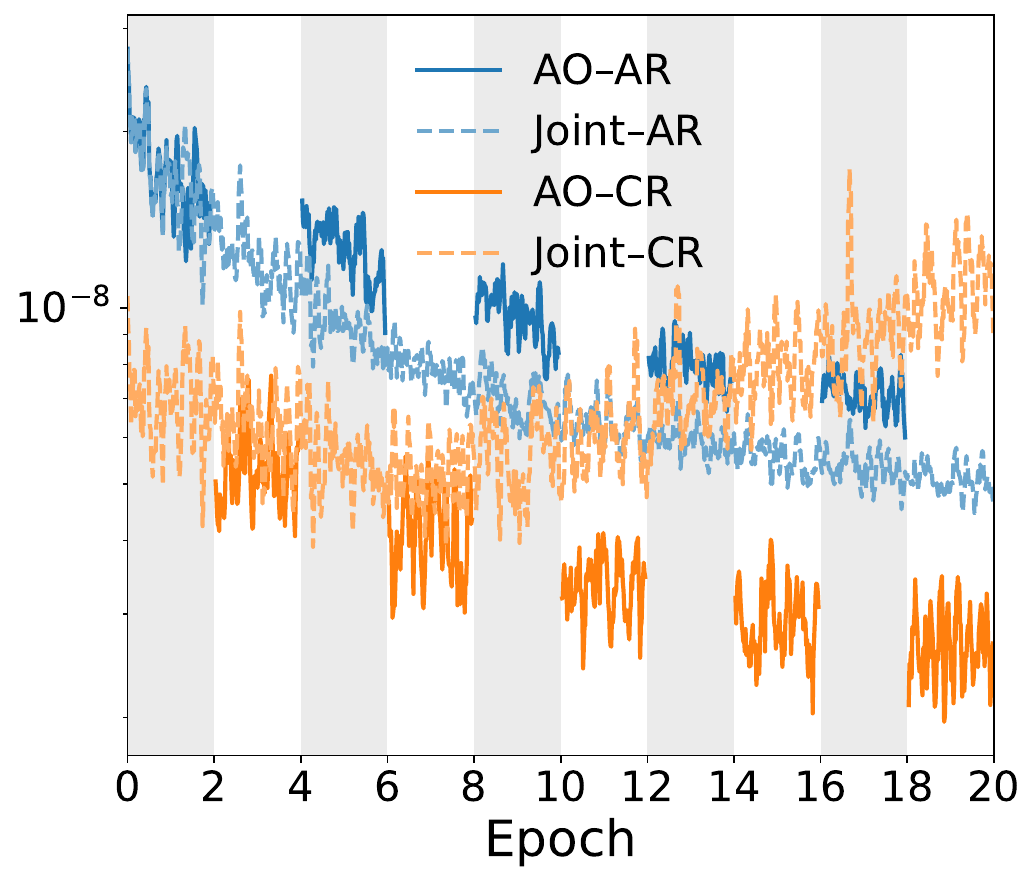}
    \caption{ETTh2}
  \end{subfigure}%
  \begin{subfigure}[t]{0.33\linewidth}
    \centering\includegraphics[width=\linewidth]{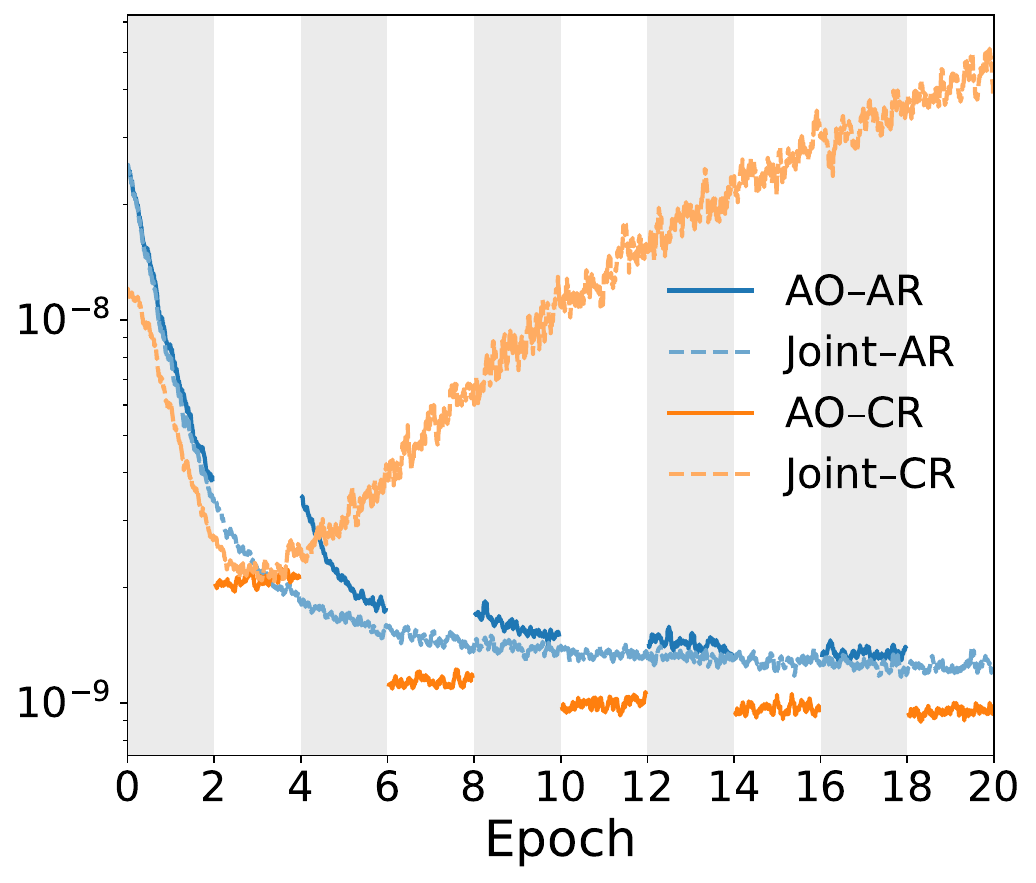}
    \caption{ETTm1}
  \end{subfigure}

  \vspace{0.5em}

  % Row 3 (single)
  \begin{subfigure}[t]{0.33\linewidth}\centering\mbox{}\end{subfigure}%
  \begin{subfigure}[t]{0.33\linewidth}
    \centering\includegraphics[width=\linewidth]{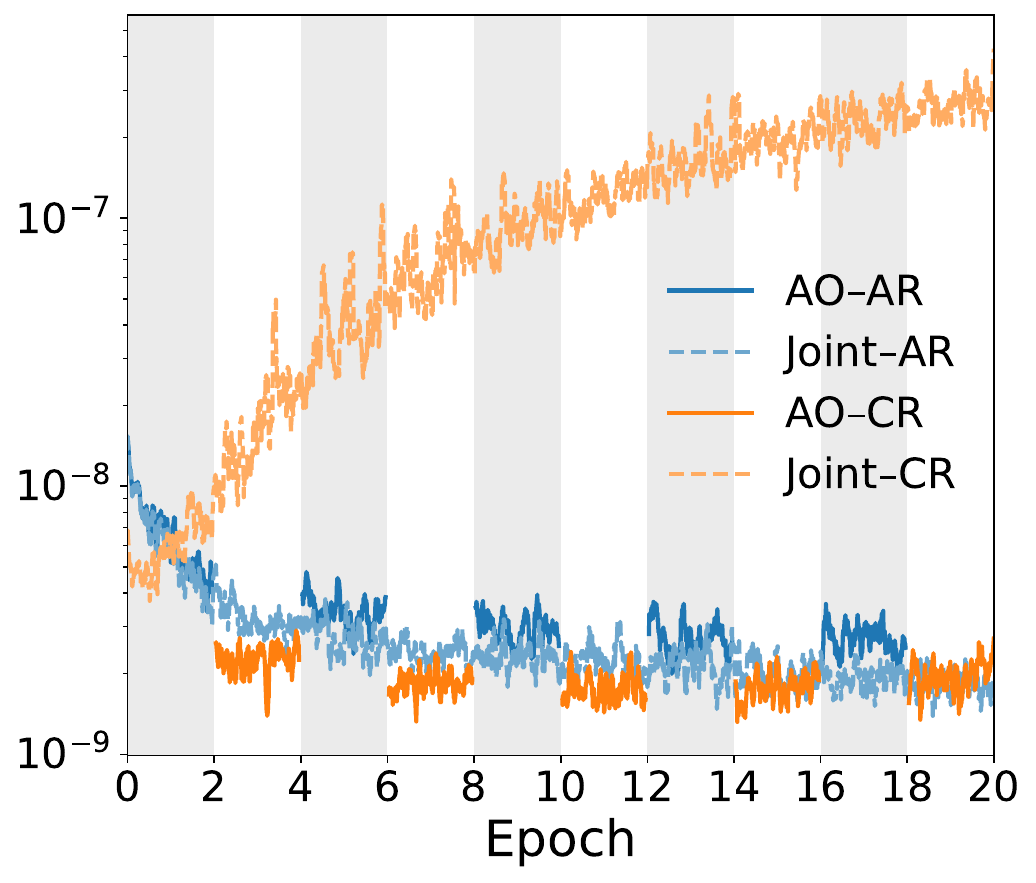}
    \caption{ETTm2}
  \end{subfigure}%
  \begin{subfigure}[t]{0.33\linewidth}\centering\mbox{}\end{subfigure}

  \caption{\textbf{Prediction length $=336$.} Variance of AR/CR gradients under joint training across seven datasets.}
  \label{fig:gv-7-336}
\end{figure}

% -------------------------------------------------------
\newpage
\subsection{Prediction length = 720}
\begin{figure}[H]
  \centering
  % Row 1
  \begin{subfigure}[t]{0.33\linewidth}
    \centering\includegraphics[width=\linewidth]{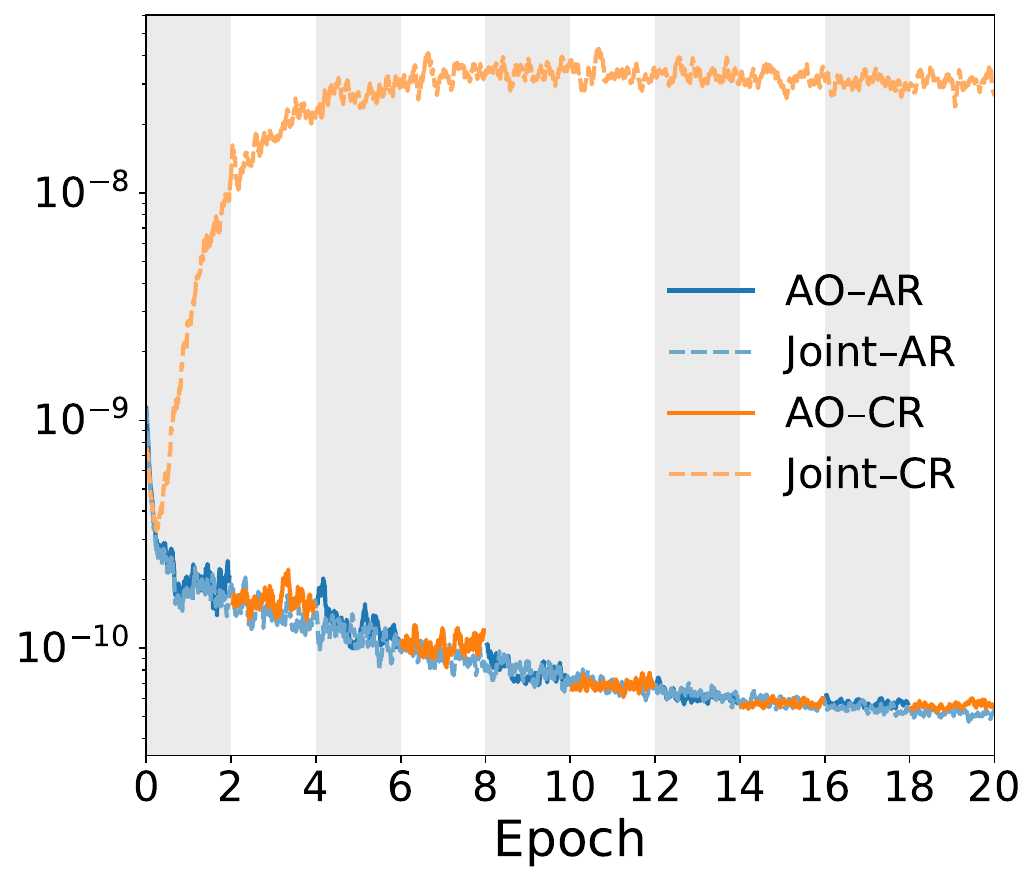}
    \caption{Weather}
  \end{subfigure}%
  \begin{subfigure}[t]{0.33\linewidth}
    \centering\includegraphics[width=\linewidth]{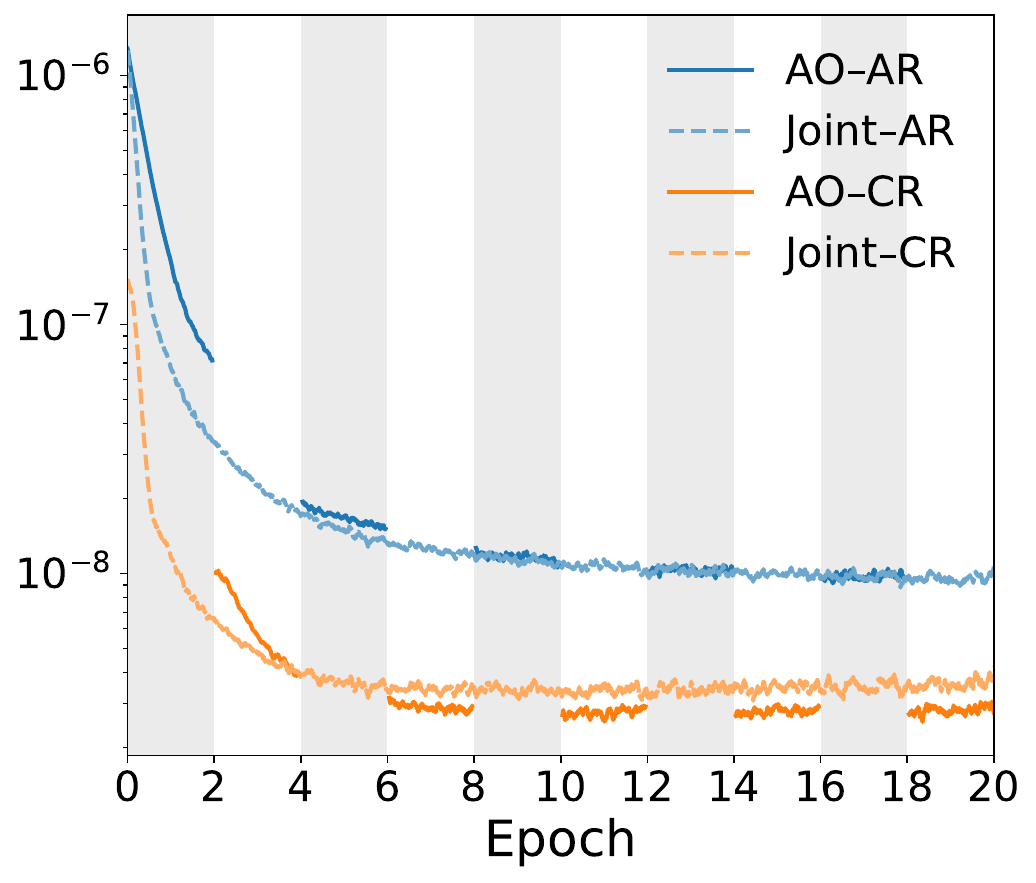}
    \caption{Traffic}
  \end{subfigure}%
  \begin{subfigure}[t]{0.33\linewidth}
    \centering\includegraphics[width=\linewidth]{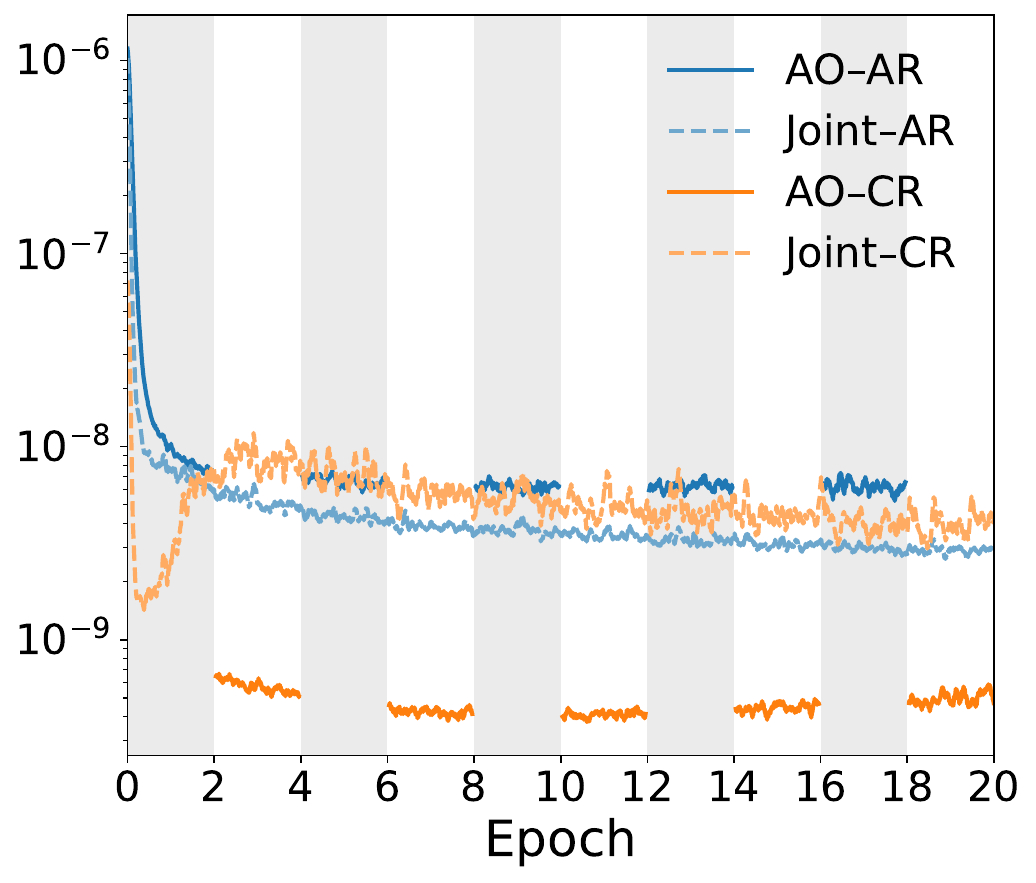}
    \caption{Electricity}
  \end{subfigure}

  \vspace{0.5em}

  % Row 2
  \begin{subfigure}[t]{0.33\linewidth}
    \centering\includegraphics[width=\linewidth]{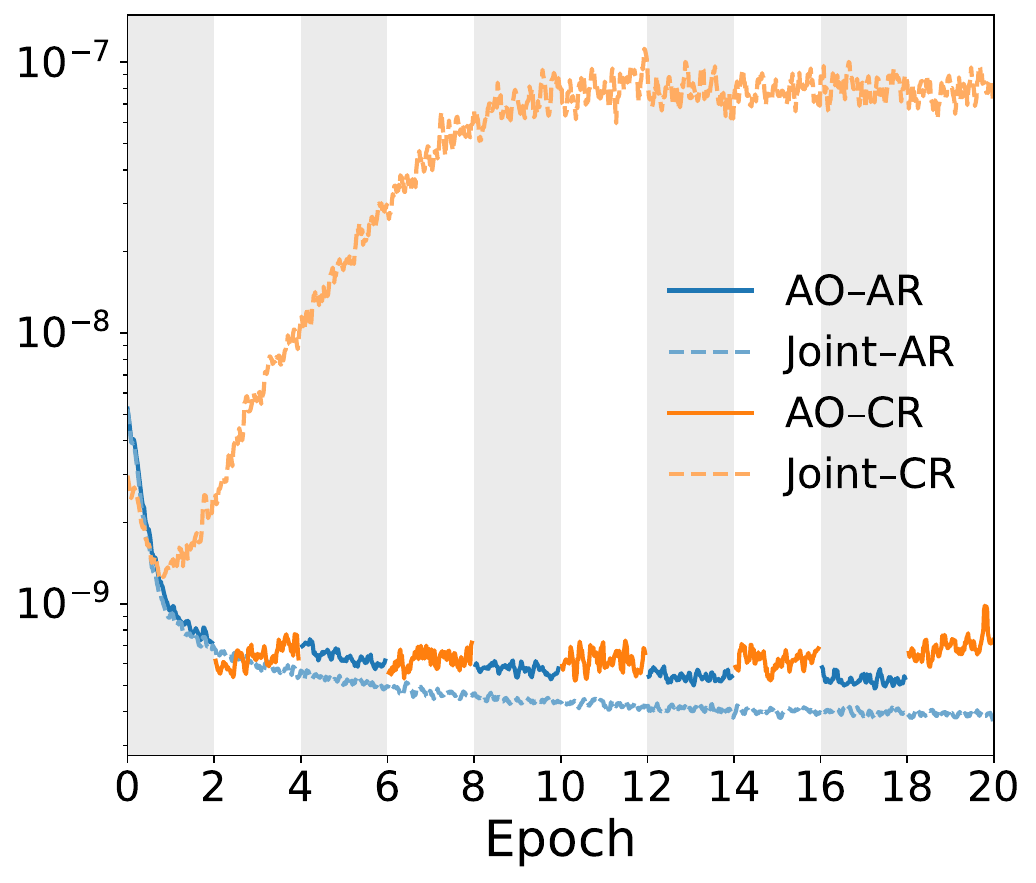}
    \caption{ETTh1}
  \end{subfigure}%
  \begin{subfigure}[t]{0.33\linewidth}
    \centering\includegraphics[width=\linewidth]{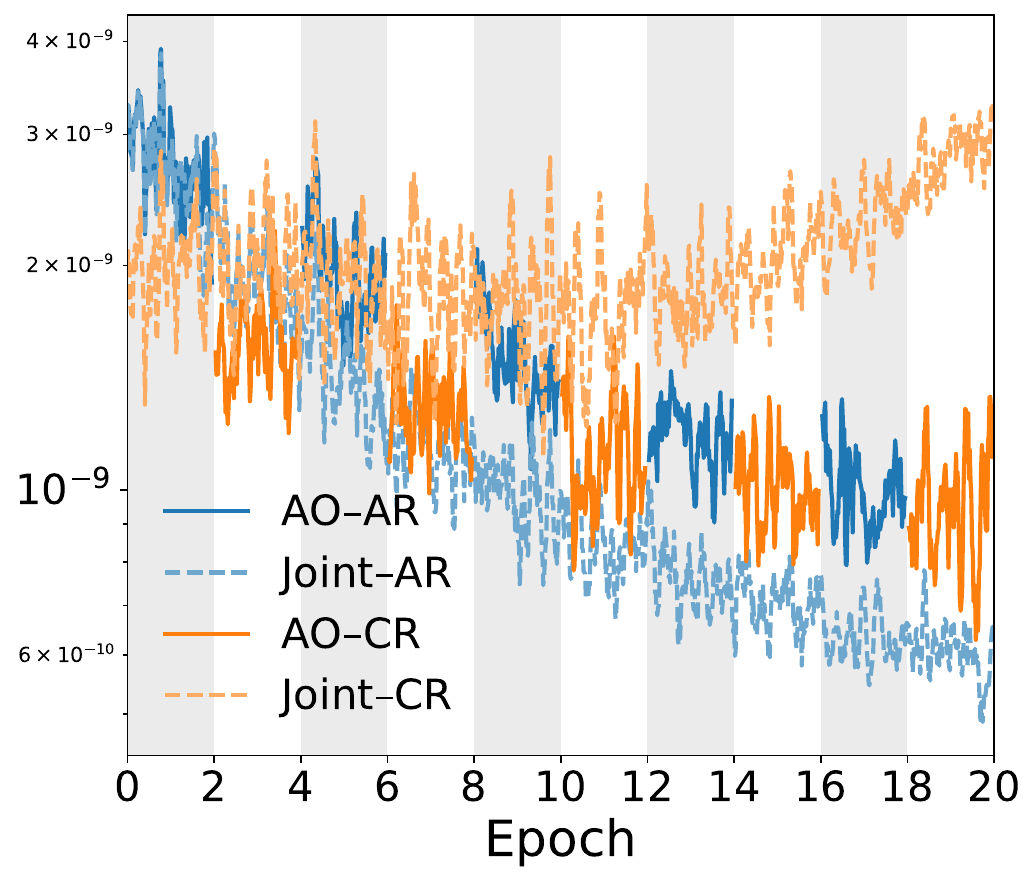}
    \caption{ETTh2}
  \end{subfigure}%
  \begin{subfigure}[t]{0.33\linewidth}
    \centering\includegraphics[width=\linewidth]{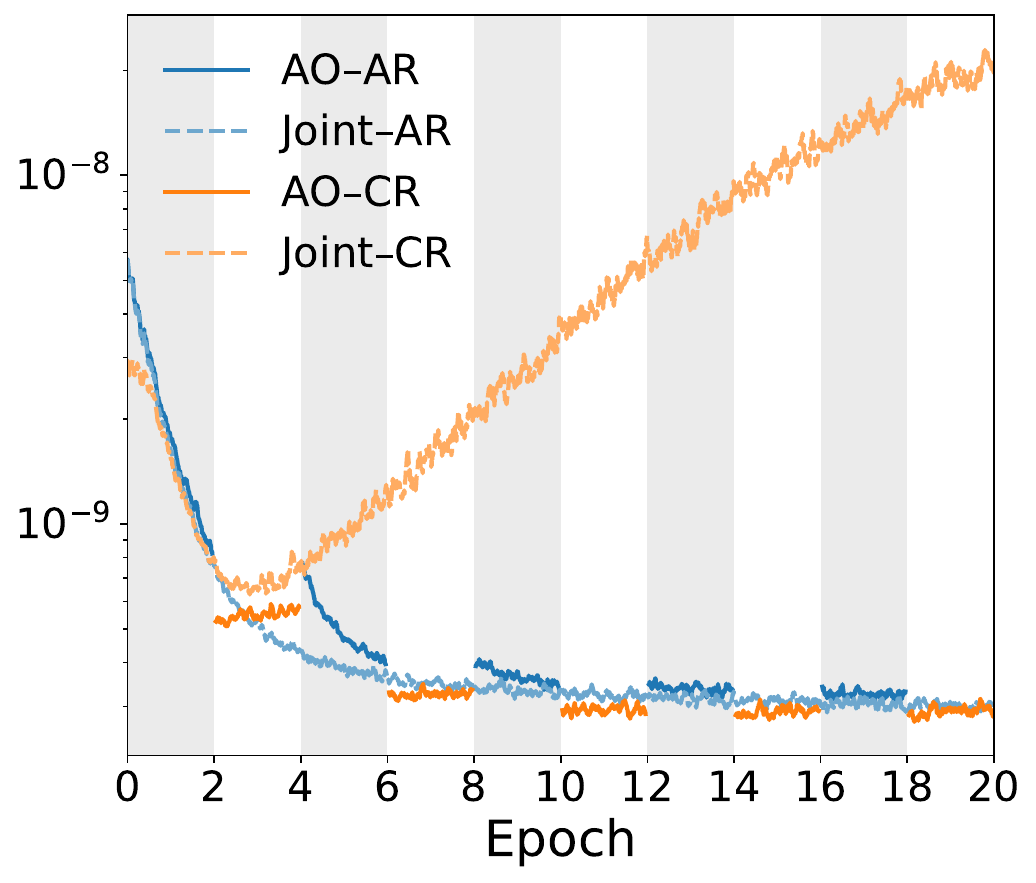}
    \caption{ETTm1}
  \end{subfigure}

  \vspace{0.5em}

  % Row 3 (single)
  \begin{subfigure}[t]{0.33\linewidth}\centering\mbox{}\end{subfigure}%
  \begin{subfigure}[t]{0.33\linewidth}
    \centering\includegraphics[width=\linewidth]{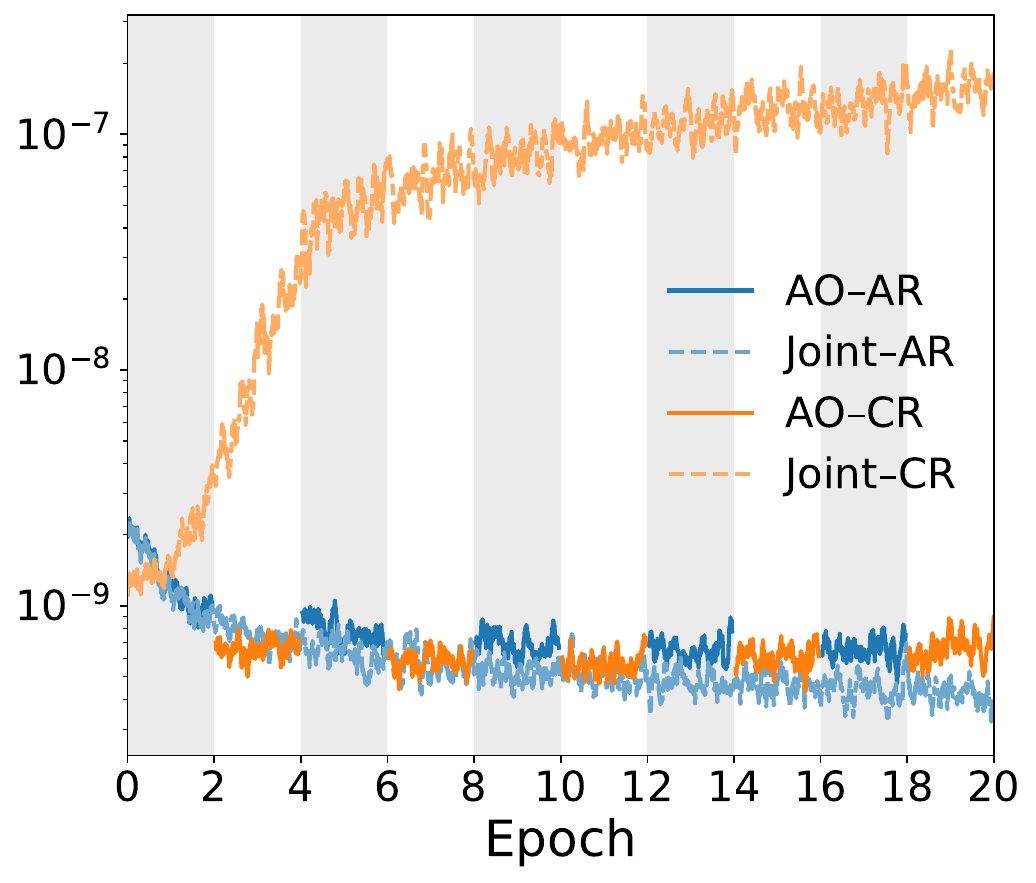}
    \caption{ETTm2}
  \end{subfigure}%
  \begin{subfigure}[t]{0.33\linewidth}\centering\mbox{}\end{subfigure}

  \caption{\textbf{Prediction length $=720$.} Variance of AR/CR gradients under joint training across seven datasets.}
  \label{fig:gv-7-720}
\end{figure}

\end{document}